\pdfoutput=1

\documentclass[11pt]{article}

\usepackage[table]{xcolor}

\usepackage{arabtex}
\usepackage{utf8}
\usepackage[utf8]{inputenc}

\usepackage[final]{acl}

\usepackage{amsmath}

\usepackage{newtxtext}
\usepackage{latexsym}
\usepackage{appendix}
\usepackage[T1]{fontenc}

\usepackage{microtype}

\usepackage{inconsolata}

\usepackage[breakable]{tcolorbox}
\usepackage{graphicx}

\usepackage{amssymb}
\usepackage{pifont}
\newcommand{\cmark}{\textcolor{green!40!black}{\ding{51}}}
\newcommand{\xmark}{\textcolor{red}{\ding{55}}}

\usepackage[stable]{footmisc}
\usepackage{hyperref}
\usepackage{todonotes}
\usepackage{arydshln} 
\usepackage{adjustbox} 

\usepackage{booktabs}

\usepackage{cuted} 
\usepackage{tikz}

\usepackage{array}              
\usepackage{caption}            
\usepackage{enumitem}

\setlist[itemize]{noitemsep, topsep=0pt, parsep=0pt, partopsep=0pt, leftmargin=*}
\usepackage{makecell}
\usepackage{xspace}
\usepackage{multirow} 
\usepackage{soul} 
\usepackage{tikz}       
\usepackage{array} 
\newcolumntype{H}{>{\setbox0=\hbox\bgroup}c<{\egroup}@{}}

\definecolor{ptcolor}{HTML}{FFD700}  
\definecolor{ftcolor}{HTML}{17BECF}  
\newcommand{\dsdiamond}[1]{%
  \tikz[baseline=-1ex, x=0.75ex, y=0.75ex]
    \draw[fill=#1, draw=#1] 
      (0,0.7) -- (0.7,0) -- (0,-0.7) -- (-0.7,0) -- cycle;%
}

\definecolor{ftcolor}{HTML}{17BECF}  

\newcommand{\dsft}[1]{\dsdiamond{ftcolor}} 

\newcommand{\ourdataset}{\textsc{Pearl}\xspace}

\newcommand{\pearlx}{\textsc{Pearl-X}\xspace}

\newcommand{\pearlt}{\textsc{Pearl-Lite}\xspace}

\newcommand{\annotators}{37\xspace}

\usepackage{placeins}  
\usepackage{capt-of} 
\usepackage{caption}   
\usepackage{cuted} 

\title{
\raisebox{-2.1ex}{\protect\includegraphics[height=3.99\fontcharht\font`\B]{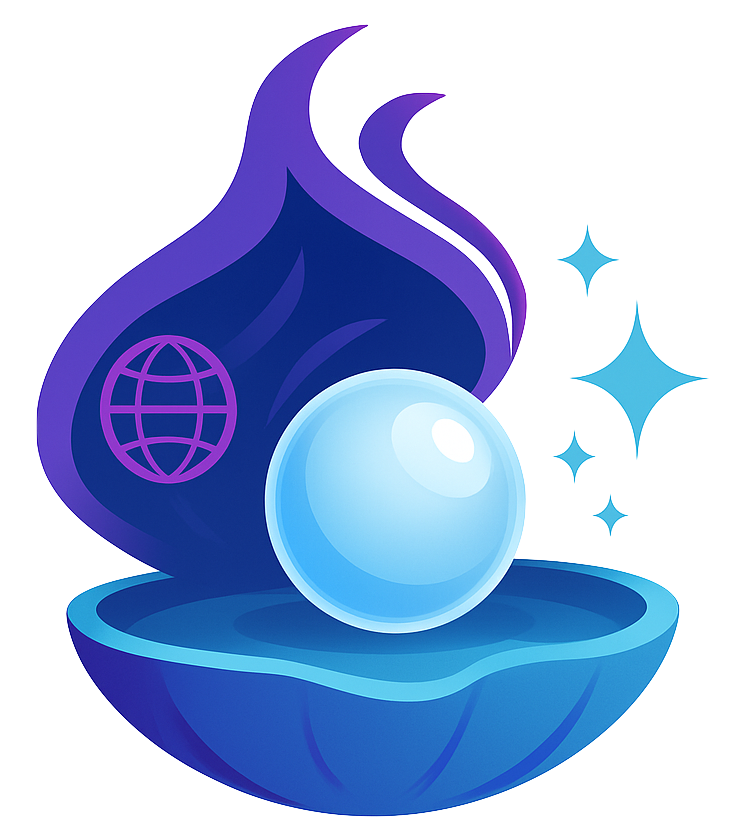}} Pearl: A Multimodal Culturally-Aware Arabic Instruction Dataset}

\author{
\begin{minipage}[t]{\textwidth}
\centering
\normalfont
Fakhraddin Alwajih\textsuperscript{1},
        Samar M. Magdy\textsuperscript{1},
        Abdellah El Mekki\textsuperscript{1},
        Omer Nacar\textsuperscript{2,3},
        Youssef Nafea\textsuperscript{4},
        Safaa Taher Abdelfadil\textsuperscript{4},
        Abdulfattah Mohammed Yahya\textsuperscript{5,6},
        Hamzah Luqman\textsuperscript{7},
        Nada Almarwani\textsuperscript{8},
        Samah Aloufi\textsuperscript{8},
        Baraah Qawasmeh\textsuperscript{9},
        Houdaifa Atou\textsuperscript{10},
        Serry Sibaee\textsuperscript{2},
        Hamzah A. Alsayadi\textsuperscript{11},
        Walid Al-Dhabyani\textsuperscript{12,5},
        Maged S.\ Al-shaibani\textsuperscript{7},
        Aya El Aatar \textsuperscript{13},
        Nour Qandos\textsuperscript{14},
        Rahaf Alhamouri\textsuperscript{15},
        Samar Ahmad\textsuperscript{16},
        Mohammed Anwar Al-Ghrawi\textsuperscript{17},
        Aminetou Yacoub\textsuperscript{18},
        Ruwa AbuHweidi\textsuperscript{19},
        Vatimetou Mohamed Lemin\textsuperscript{18},
        Reem Abdel-Salam\textsuperscript{12},
        Ahlam Bashiti \textsuperscript{19},
        Aisha Alansari\textsuperscript{7},
        Ahmed Ashraf\textsuperscript{7},
        Nora Alturayeif\textsuperscript{20},
        Alcides Alcoba Inciarte\textsuperscript{1},
        Adel Ammar\textsuperscript{2},
        Abdelrahim A.\ Elmadany\textsuperscript{1},
        Mohamedou Cheikh Tourad\textsuperscript{18},
        Ismail Berrada\textsuperscript{10},
        Mustafa Jarrar\textsuperscript{21,19},
        Shady Shehata\textsuperscript{22}, 
        Muhammad Abdul-Mageed\textsuperscript{1,22} \\[1em]
{
\textsuperscript{1} The University of British Columbia,
            \textsuperscript{2} Prince Sultan University,
            \textsuperscript{3} Tuwaiq Academy,
            \textsuperscript{4} Independent Researcher,
            \textsuperscript{5} Hadhramout University,
            \textsuperscript{6} Misr University for Science \& Technology,
            \textsuperscript{7} KFUPM,
            \textsuperscript{8} Taibah University,
            \textsuperscript{9} WMU,
            \textsuperscript{10} UM6P,
            \textsuperscript{11} IBB University,
            \textsuperscript{12} Cairo University,
            \textsuperscript{13} UCA,
            \textsuperscript{14} Qafza Tech,
            \textsuperscript{15} JUST,
            \textsuperscript{16} KAUST,
            \textsuperscript{17} Damascus University,
            \textsuperscript{18} University of Nouakchott,
            \textsuperscript{19} Birzeit University,
            \textsuperscript{20} Imam Abdulrahman Bin Faisal University, \\
            \textsuperscript{21} Hamad Bin Khalifa University, 
            \textsuperscript{22} Invertible AI
}\\[0.5em]
\texttt{\normalsize \{fakhr.alwajih, muhammad.mageed\}@ubc.ca}
\end{minipage}
}

\setcode{utf8}

\begin{document}

\maketitle



\leavevmode\vspace{270pt}

\begin{strip}
\centering
\includegraphics[width=0.82\textwidth]{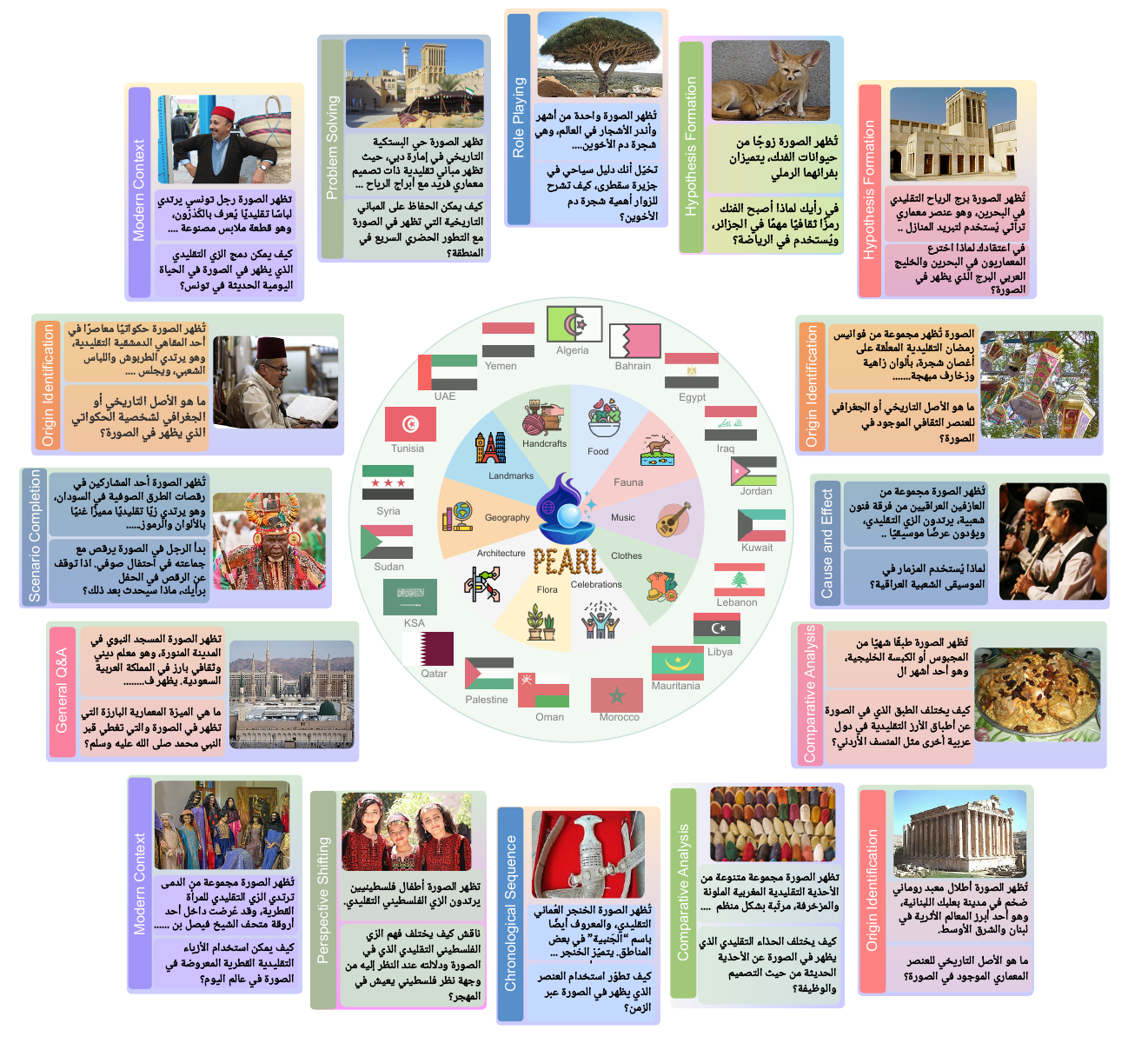}
\captionof{figure}{Overview of \ourdataset, covering all but three Arab countries (16 illustrated here) and presenting representative examples from 11 of our 13 challenging question categories. These questions require reasoning and deep cultural knowledge (English translations provided in Appendix~\ref{appdx_table:examples}). Prompts span diverse cultural domains, with images sourced from Wikipedia and other publicly available resources.}
\label{fig:main}
\end{strip}

\twocolumn

\begin{abstract}
Mainstream large vision-language models (LVLMs) inherently encode cultural biases, highlighting the need for diverse multimodal datasets. To address this gap, we introduce \ourdataset, a large-scale Arabic multimodal dataset and benchmark explicitly designed for cultural understanding. Constructed through advanced agentic workflows and extensive human-in-the-loop annotations by \annotators annotators from across the Arab world, \ourdataset comprises over $309$K multimodal examples spanning ten culturally significant domains covering all Arab countries. We further provide two robust evaluation benchmarks (\ourdataset and \pearlt) along with a specialized subset (\pearlx) explicitly developed to assess nuanced cultural variations. Comprehensive evaluations on state-of-the-art open and proprietary LVLMs demonstrate that reasoning-centric instruction alignment substantially improves models’ cultural grounding compared to conventional scaling methods. \ourdataset~establishes a foundational resource for advancing culturally-informed multimodal modeling research. All datasets and benchmarks are publicly available.\footnote{\href{https://github.com/UBC-NLP/pearl}{https://github.com/UBC-NLP/pearl}}
\end{abstract}
\section{Introduction}

Mainstream large vision-language models (LVLMs) predominantly encode Western perspectives, often neglecting diverse cultural traditions. Recent efforts have partially addressed this issue through multilingual datasets~\cite{romero2024cvqa,liu2025culturevlm} and benchmarks like CultureVLM~\cite{liu2025culturevlm}, GIMMICK~\cite{schneider2025gimmick}, CVQA~\cite{romero2024cvqa}, and related initiatives~\cite{vayani2024all}. However, some of these resources lack sufficient cultural depth, often relying on translated content, limited image sets, narrow topical coverage, and simplistic factual questions unsuitable for evaluating advanced LVLM capabilities. Furthermore, they frequently assume local cultural homogeneity, overlooking nuanced regional variations even within shared cultural artifacts. Thus, developing high-quality, culturally nuanced multimodal resources that adequately challenge contemporary LVLMs remains significantly underexplored, especially for many global regions, including the Arab world.

Addressing this gap, we introduce \ourdataset, a large-scale Arabic multimodal instruction dataset and benchmark explicitly designed for cultural understanding. To ground our work, we operationalize \textit{culture} through the lens of \textit{cultural heritage}. That is, the focus of the work is on the enduring traditions, values, and practices validated by native speakers~\cite{pawar2025survey}. This framing provides a clear scope that distinguishes deeply rooted customs from contemporary, globalized practices. Our approach aligns with recent efforts to systematically measure and model cultural dimensions in language models, which highlight the importance of heritage and regional specificity~\cite{pawar2025survey,liu2025culturally}.

\ourdataset is the outcome of an extensive collaborative effort involving a diverse community of \annotators contributors, all of whom are authors of this work, spanning the Arab region. By leveraging advanced agentic workflows alongside iterative human-in-the-loop refinement, our dataset integrates sophisticated LLM and LVLM outputs with the nuanced cultural expertise of native annotators. Covering ten culturally significant domains, \ourdataset~authentically captures Arab cultural heritage. Moreover, we structure it around ten distinct question categories designed to test sophisticated LVLM capabilities, including \textit{hypothesis formation, problem-solving, comparative analysis}, and \textit{chronological sequencing}.

To specifically evaluate subtle cultural variations overlooked by existing benchmarks, we also introduce \pearlx(Figure~\ref{fig:shared_workflow}, a novel benchmark highlighting culturally shared yet visually distinct concepts (e.g., \textit{coffee}) across different Arab contexts. Unique among benchmarks, \pearlx incorporates both text-to-single-image and text-to-multiple-image pairings, enabling richer assessments of LVLM performance on complex multimodal tasks. We leverage our benchmarks to systematically evaluate a range of open and proprietary LVLMs across diverse sizes and capabilities. Our results demonstrate that models incorporating reasoning-based cultural alignment substantially outperform those relying solely on conventional scaling approaches.

The rest of this paper is organized as follows: Section~\ref{sec:related} reviews related work on multilingual and cultural benchmarks. Section~\ref{sec:method} details our data collection, annotation pipeline, and benchmark design. Section~\ref{sec:experiments} describes the evaluation protocol and experimental setup. Section~\ref{sec:results} presents our findings and analysis. Section~\ref{sec:pearlx} introduces the novel \pearlx benchmark .Finally, Section~\ref{sec:conclusion} concludes the paper and outlines future directions.

\section{Related Work}
\label{sec:related}

\paragraph{Multilingual VQA Datasets.} Multilingual VQA datasets are predominantly constructed by generating QA pairs in various target languages, often alongside English counterparts. The creation methodologies for these datasets include manual annotation \cite{romero2024cvqa, liu2021visually, pfeiffer2021xgqa, das2024exams}, fully automatic generation \cite{becattini2023viscounth}, or automatic translation followed by human verification \cite{changpinyo2022maxm, vayani2024all, wang2024m4u, das2024exams, liu2025culturevlm}. Notable examples include VISCOUNTH \cite{becattini2023viscounth}, a large-scale Italian-English dataset featuring $500$K images and $6.5$M QA pairs that were semi-automatically generated using an existing ontology-based knowledge graph. Another significant resource is the M5 benchmark \cite{schneider2024m5}, which encompasses eight datasets across five vision tasks (introducing M5-VGR and M5-VLOD). While M5 spans $13$ languages, its collection of $79,470$ images presents a limitation considering its broad linguistic coverage.

\paragraph{Multi-Cultural VQA Datasets.} Relatively few VQA datasets prioritize cultural diversity in their image curation and QA generation processes. For instance, CulturalVQA \cite{nayak2024benchmarking} incorporates images representing cultural concepts ($2,378$ image-QA pairs, $11$ countries) but is limited to English. The ALM-bench \cite{vayani2024all} benchmark covers cultural aspects from $73$ countries and $100$ languages ($2,929$ images, $22,763$ QA pairs, $13$ domains). Its QA pairs were automatically generated and translated using GPT-$4$o, followed by human verification. MaXM \cite{changpinyo2022maxm} is a test-only VQA dataset in seven languages (five scripts), featuring $2,142$ auto-generated and human-verified QA pairs. CVQA \cite{romero2024cvqa} provides QA pairs in the local languages of $30$ countries, alongside English translations for $5,239$ images ($10,374$ questions). More extensively, CultureVerse \cite{liu2025culturevlm} contains $74,959$ images covering $19,682$ cultural concepts from $188$ countries, with its GPT-$4$o- generated QA pairs validated through automated and manual checks.

\paragraph{Arabic Cultural VQA.} Despite the growing interest in VQA, Arabic cultural representation in existing datasets remains sparse. The Henna dataset \cite{alwajih2024peacock} is specifically dedicated to Arabic culture, comprising $1,132$ images from $11$ Arabic countries. Other datasets offer minimal Arabic content: ALM-bench \cite{vayani2024all} includes $168$ images from Saudi Arabia, UAE, and Egypt; CVQA \cite{romero2024cvqa} contains approximately $200$ Egyptian images; and CultureVerse \cite{liu2025culturevlm} features only seven Libyan and 272 Egyptian images. More recently, JEEM \cite{kadaoui2025jeem} introduced a benchmark for five Arabic dialects, consisting of $10.89$K QA pairs and $2,178$ images across $13$ cultural domains. While JEEM improves Arabic VQA coverage, it is limited to four question categories and highlights the ongoing need for broader and deeper cultural representation.

To the best of our knowledge, \ourdataset is the first large-scale, culturally diverse Arabic multimodal benchmark carefully constructed through extensive human supervision, covering a wide range of cultural domains and challenging question types requiring reasoning and deep cultural knowledge. A comparative analysis with existing datasets is presented in Table~\ref{tab:comparison_related_work}. Additional discussion on cultural biases in LVLMs and existing monolingual VQA datasets is provided in Appendix~\ref{app:related_work}.

\begin{table*}[]
\resizebox{\textwidth}{!}{
\begin{tabular}{cllHllllllll}
\toprule 
\textbf{Category} & \textbf{Dataset}                                                                         &  \textbf{Lang.} & \textbf{AraDia.} & \textbf{Domains} & \textbf{Images} & \textbf{AraQA/Total} & \textbf{Q-Type}      & \textbf{Q-Form} & \textbf{Ann.} & \textbf{CC} & \textbf{BC} \\ \midrule 


\multirow{13}{*}{\rotatebox[origin=]{90}{\textbf{\colorbox{blue!10}{Multilingual}}}}  & CVQA$^{\star}$   \cite{romero2024cvqa}                      & $30$ & $1$              & $10$            & $5,239$        & $200$/$10.4$K          & MCQ                 & Fixed          &  M           & \cmark                & \xmark               \\
 & MMBench \cite{liu2025mmbench}                    & $2$             & \xmark   & $20$            & $2974$         & $00$/$3.2$K           & MCQ                 & Fixed          &  A + M      & \xmark                & \cmark               \\
 & EXAMS-V \cite{das2024exams}                     & $11$   & \xmark            & $20$            & $20,932$       & $00$/$20.9$K          & MCQ                 & Diverse        &  M           & \xmark                & \cmark               \\
 & MaRVL  \cite{liu2021visually}                                                              & $5$   & \xmark             & $1$             & $5464$         & $00$/$5,464$           & TF                  & Fixed          &  M           & \cmark                & \cmark               \\
 & M3Exam \cite{zhang2023m3exam}                                                              & $9$  & \xmark              & $3$             & $2,816$        & $00$/$12.3$K          & MCQ                 & Diverse        &  A + M      & \xmark                & \cmark               \\
 & MaXM \cite{changpinyo2022maxm}                                                             & $7$   & \xmark             & -             & $700$          & $00$/$2.1$K            & SVQA                & Fixed          &  A + M      & \cmark                & \cmark               \\
 & xGQA \cite{pfeiffer2021xgqa}                                                                & $8$    & \xmark            & -             & $459$          & $00$/$14.4$K          & Y/N, SVQA           & Fixed          &  M           & \xmark                & \xmark               \\
 & M4U \cite{wang2024m4u}                                                                    & $3$      & \xmark          & $64$            & $8,931$        & $00$/$8.9$K           & MCQ                 & Fixed          &  A + M      & \xmark                & \cmark               \\
 & CultureVerse$^{\star}$  \cite{liu2025culturevlm}            & $188$  & $2$            & $15$            & $74,959$       & $279$/$196.7$K         & MCQ, SVQA, LVQA     &     Diverse           &  A + M      &      \xmark                 & \xmark               \\
 & ALM-Bench$^{\star}$    \cite{vayani2024all}                                                                             & $100$         & $3$      & $19$           & $2,929$        & $1,008$/$22.7$K          & MCQ, SVQA, LVQA, TF & Diverse        &  A + M      & \cmark                & \cmark               \\ 
 & WorldCusine\cite{winata2024worldcuisines}                                                                                 & $30$     & \xmark           & -         & $6,045$        &  $00$/$1.2$M             &  MCQ, open-ended       & -          &  A           & \xmark                & \xmark               \\

\midrule
\multirow{6}{*}{\rotatebox[origin=]{90}{\textbf{\colorbox{orange!10}{X-Specific}}}}  & MMMU \cite{yue2024mmmu}                                                                                       & English & \xmark               & $30$            & $11,550$         &                 & MCQ, SVQA           & Fixed          &  M           & \xmark                & \xmark               \\

 & ViTextVQA \cite{vannguyen2024ViTextVQA}                                                                                      & Vietnamese     & \xmark            & -            & $16,762$         &   $00$/$50.3$K              & LVQA         & Fixed          &  M           & \xmark                & \xmark               \\
 & MMT-Bench \cite{ying2024mmtbench}                                                                                      & English    & \xmark            & -            & -         &   $00$/$31.3$K              & MCQ           & Fixed          &  A + M           & \xmark                & \xmark               \\
 & JMMMU \cite{onohara2024jmmmu}                                                                                       & Japanese     & \xmark            & -            & $1,118$        &   $00$/$1.3$K             & MCQ           & Fixed          &  A + M           & \cmark                & \xmark               \\
 & HaVQA \cite{parida-etal-2023-havqa}                                                                                     & Hausa   & \xmark             & -            & $1,555$        &   $00$/$6$K             &  LVQA          & Fixed          &  M           & \xmark                & \xmark               \\

& CULTURALVQA \cite{nayak2024benchmarking}                    & English             & \xmark   & -            & $2328$         & $2378$           & open-ended                 & Diverse          & M      & \cmark                & \cmark               \\
 & VLBiasBench\cite{wang2024vlbiasbench}                                                                                 & English      & \xmark        & $11$           & $46,848$        & $00$/$128.3$K             &   LVQA, MCQ        & -          &  A + M           & \xmark                & \cmark               \\

\midrule 
\multirow{2}{*}{\rotatebox[origin=]{0}{\textbf{\colorbox{green!10}{Arabic}}}} &              Henna \cite{alwajih2024peacock}    &     Arabic   & $11$       &  $5$               & $120$  &  $1,132$       &  SVQA    &    Diverse            &  A + M           & \cmark                & \xmark        \\
 & CAMEL-Bench \cite{ghaboura2024camel}                                                                                 & Arabic   & \xmark             & $8$           & -        &  $29$K/$29$K             &  MCQ          & Fixed          &  A + M          & \xmark                & \xmark               \\
 & JEMM \cite{kadaoui2025jeem}                                                                                 & Arabic   & $4$             & $13$           & $2,178$        &  $10,890$K             &  open-ended          & Diverse          &  A           & \cmark                & \cmark               \\ \cmidrule{2-12}

 &             \includegraphics[scale=0.013]{figures/pearl_logo_1.png} Pearl (ours)               &     Arabic  & \xmark     & $10$   & $12$k           &$309$k/$309$k   & 13-Q-Type\textsuperscript{$\star\star$} & Diverse        &  A + M      & \cmark                & \cmark      \\
\bottomrule 
\end{tabular}%
}
\caption{
Comparison of related visual datasets, covering multilingual, cross-specific, and Arabic resources. \textbf{Lang.}:  number of languages for multilingual datasets or the language name for cross-specific language datasets. \textbf{AraQA/Total}: number of Arabic questions compared to the total number of covered questions. \textbf{Q-Type}: types of questions, for instance, long VQA (LVQA), short VQA (SVQA), and True False (TF) questions. \textbf{Q-Form}: question phrasing of each image.
\textbf{Ann.}: annotation method used while creating the datasets (``\texttt{M}:'' manual data collection, filtering, and annotation;  ``\texttt{A}:'' automatic). \textbf{CC}: inclusion of cultural content. \textbf{BC}: use of bias correction. \textsuperscript{$\star$}The CVQA contains Arabic samples. \textsuperscript{$\star\star$}\ourdataset has 13 different Q-types, as described in Table~\ref{tab:cultural_distribution}.}
\label{tab:comparison_related_work}
\end{table*}

\section{Methodology}
\label{sec:method}
\begin{figure*}[t]
    \centering
    \includegraphics[width=1.00\linewidth]{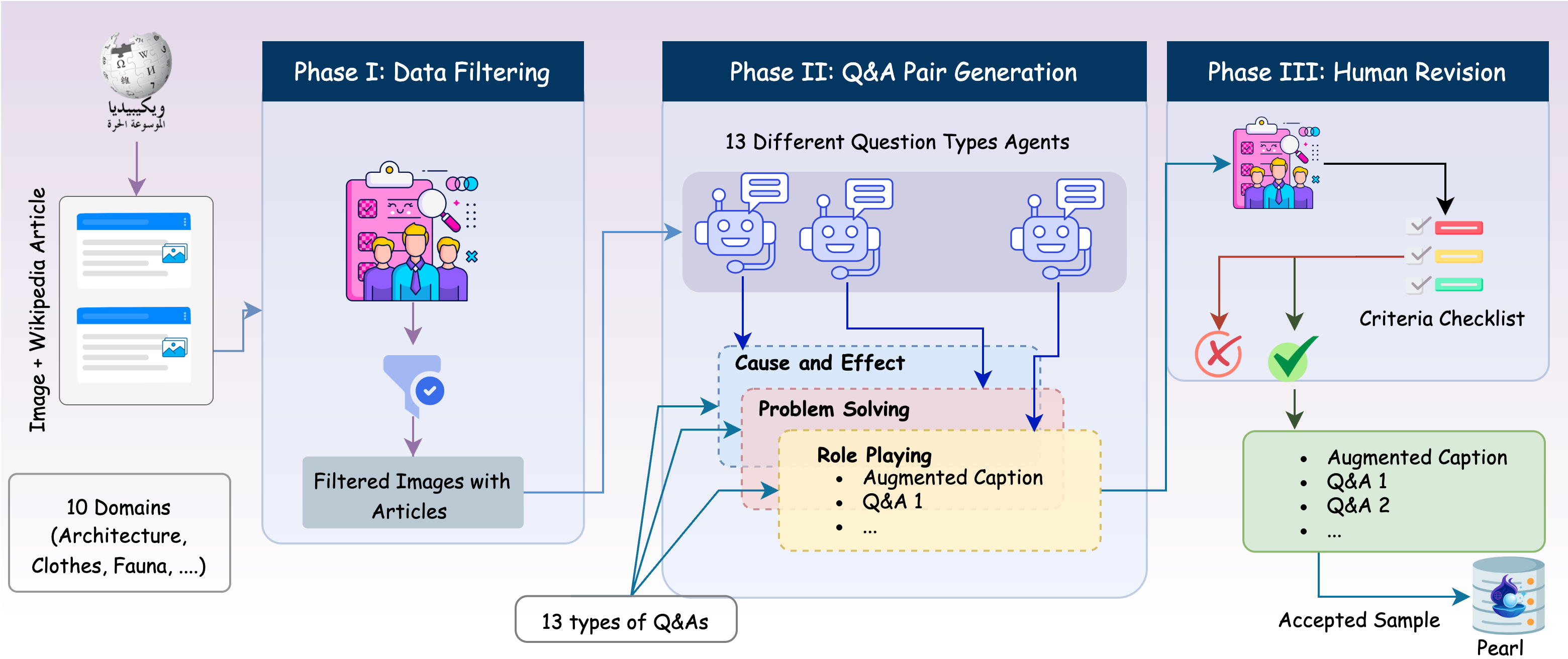}
    \caption{Illustrates the three main stages of our data annotation pipeline. (1) selecting culturally relevant images from \textit{Wikipedia} articles across ten predefined domains (e.g., food, music, architecture). (2) Agents generate augmented captions and Q\&A pairs across $13$ question types. (3) Human reviewers apply a quality checklist to accept or reject samples for the final dataset.}
    \label{fig:workflow}
\end{figure*}

\subsection{Annotation Process}

\textbf{Annotation team.} Our pipeline for \ourdataset involves the use of an agentic workflow intertwined with human effort by our annotation team. The team comprises \annotators local members from nine Arab countries\footnote{Egypt, Palestine, Yemen, Morocco, Tunisia, Syria, Jordan, Saudi Arabia, Mauritania} all native Arabic speakers, each holding at least a bachelor's degree. We cover all Arab countries except three\footnote{These are Comoros, Djibouti, and Somalia.} For each of these countries, we assigned at least two members either from the same country or a neighboring country (to ensure familiarity with local culture). 

\noindent \textbf{Annotation Guidelines.} Over a period of six months, we developed an extensive set of annotation guidelines covering the different stages of the project. This includes a number of dimensions: (i) introduction of cultural domains and criteria of data selection based on uniqueness to culture and relevance of images to the articles we collect, (ii) illustrated definitions of question types and criteria for characterizing high-quality questions and answers and how to improve these, and (iii) illustrative screenshots of the annotation platform itself. Our full annotation guidelines are available at 
~\href{https://github.com/UBC-NLP/pearl}{https://github.com/UBC-NLP/pearl}

\noindent \textbf{Annotation Platform and Communication.} We utilized Label Studio~\cite{tkachenko2020label} as our primary annotation platform, organizing annotators into country-specific teams. Each annotator carefully reviewed content relevant to their respective country, closely following our detailed guidelines. To ensure effective coordination, we maintained a dedicated Slack channel for real-time communication, feedback, and progress updates. Additionally, we conducted weekly full-team meetings, recorded and distributed to all team members, complemented by smaller team meetings scheduled as needed. Annotators received recorded video tutorials demonstrating practical annotation examples and addressing common challenges. Further details regarding the annotation process and platform setup can be found in Appendix~\ref{app:guidelines}.

\subsection{Human-in-the-Loop Agentic Workflow}

Our workflow begins by collecting image-article pairs from Wikipedia. Each image then undergoes a human review process to filter out irrelevant content (\textit{Phase I}). Next, we employ specialized LLM agents to generate diverse categories of questions based on the selected images (\textit{Phase II}). Finally, these image-question pairs undergo further human revision to ensure quality and cultural relevance (\textit{Phase III}). This workflow is illustrated in Figure~\ref{fig:workflow}, and we describe each phase in greater detail below.

\subsubsection{Phase I: Data Filtering}

We initiate our data collection process by selecting ten culturally significant domains: \textit{{architecture, clothing, fauna, festivals and celebrations, flora, foods, geography, handcrafts, landmarks, and music}}. These domains capture the diverse traditions and cultural identities prevalent across the Arab world, forming a robust foundation for curating a high-quality multimodal dataset comprising culturally contextualized text and images. Arabic \textit{Wikipedia} serves as our primary source due to its extensive and accessible country-specific content available in the native language. We systematically gathered relevant articles along with their corresponding images, prioritizing those that distinctly highlight culturally meaningful topics from various Arab countries. Each article was carefully categorized into one of the predefined cultural domains, guided closely by \textit{Wikipedia’s} established internal taxonomy.

Due to the heterogeneous nature of \textit{Wikipedia} content, initial retrieval yielded varied quality and cultural relevance. Hence, we implemented a hybrid human and automatic filtering strategy. First, an \textit{\textbf{automated filtering}} evaluated each article and image based on basic metadata alignment criteria, such as consistency between image captions, articles, and categorical tags. This automated filter significantly reduced noise by discarding obviously irrelevant or misaligned images. After the initial automated filtering, human annotators carefully reviewed each item using a structured annotation interface on Label Studio. During this \textit{\textbf{manual review}} phase, annotators ensured that images and articles were correctly aligned, culturally relevant, authentic, and factually accurate. Any image or article not meeting these standards of authenticity and relevance was excluded from further processing.

\subsubsection{Phase II: Q\&A Pair  Generation} \label{sec:phase2}
In this stage, filtered image-article pairs undergo an automated content generation process. We employ agents backboned with an LVLM\,---specifically \texttt{Qwen2.5-VL-72B-Instruct}---to create synthetic augmented captions and structured question--answer (Q\&A) pairs for each image. We use a fixed schema of $13$ predefined question types, each specifically tailored to assess different levels of understanding and reasoning skills. For example, we include questions involving \textit{cause-and-effect, chronological ordering, comparative analysis, modern context interpretation, hypothesis formation}, and \textit{scenario completion}. Table \ref{tab:question_types} shows the full set of our question categories, and Figure \ref{fig:main} illustrates an example for each type.

For each filtered image-article pair, the LVLM generates an augmented caption that integrates visual details from the image with relevant cultural or historical context from the associated \textit{Wikipedia} article (prompt details provided in Appendix~\ref{appdx_fig:prompt_step1}). Subsequently, the LVLM produces one augmented caption along with $2$–$4$ structured Q\&A pairs aligned explicitly with predefined reasoning categories (e.g., cause-and-effect reasoning). Each Q\&A generation prompt clearly specifies the targeted reasoning type, ensuring the resulting questions are culturally meaningful and require synthesizing both visual and textual inputs. Detailed prompts used for generating Q\&A pairs are provided in Appendices~\ref{appdx_fig:prompt_step2/part1} and \ref{appdx_fig:prompt_step2/part2}.

\subsubsection{Phase III: Human Revision}

We employ a \textit{two-stage} human-in-the-loop quality control process. In \textit{stage one}, annotators validate the augmented captions against the original image and article metadata based on four main criteria: (i) \textit{cultural authenticity,} to ensure the captions genuinely reflect the cultural context depicted in the images; (ii) \textit{visual relevance,} to confirm that the images clearly match the caption; (iii) \textit{clarity and precision,} to guarantee question-answer pairs are understandable and grammatically correct; and (iv) \textit{content accuracy,} to guarantee the question-answer pairs are consistent with the provided articles or other credible sources annotators can locate online. Figure~\ref{fig:platform_example} illustrates caption-image validation and Figure~\ref{fig:platform_step2} illustrates the question-answer review process.

In \textit{stage two} of human revision, carried out through a custom-built user interface, we sample a total of $11,000$ question-answer pairs from stage one and examine them to verify cultural relevance and ensure quality. All pairs not adhering to our revision criteria in \textit{Phase III} are rejected. Figure~\ref{fig:benchmark_revision_ui} shows a snapshot of the user interface we use for the second round of data revision.

\section{Our Datasets}
Apart from our main benchmarks (see Sections \S\ref{sec:peral_benchmark} and \S\ref{sec:pearlx}), our pipeline produces several high-quality datasets suitable for various purposes: (1) \textbf{Culturally-Relevant Images}, comprising $12,637$ manually selected images reflecting distinct country-level cultural nuances across ten domains; (2) \textbf{Augmented Captions}, a set of $135k$ carefully crafted captions via agents derived from Arabic Wikipedia articles. These enriched captions serve as a visual knowledge base, providing contextually rich and interconnected information beyond standard image descriptions, and are publicly released for community use; (3) \textbf{Automated Q\&A Pairs}, including $309$K question-answer pairs systematically generated by a suite of 13 specialized agents tailored to specific question categories; and (4) \textbf{Human-Revised Q\&A Pairs}, a high-quality subset of $16$K question-answer pairs reviewed by human annotators to ensure cultural accuracy and relevance. Collectively, these resources provide robust datasets for work involving VLMs. We now introduce our evaluation benchmarks.

\subsection{\ourdataset Benchmarking Data}\label{sec:peral_benchmark}

\textbf{\ourdataset Benchmark.} The \ourdataset benchmark is composed of~$6,301$ high-quality Q\&A pairs, carefully selected through a rigorous two-stage human evaluation process from an initial set of $16$K pairs drawn from our larger corpus of $309$K pairs. Specifically designed for model evaluation, \ourdataset includes $4,832$ \textit{closed-form} multiple-choice and True/False questions focused on recognition and interpretation of culturally relevant visual information. This component evaluates LVLMs' fundamental accuracy and factual correctness, providing a reliable baseline to assess their foundational Arabic cultural knowledge. Additionally, the benchmark contains $1,469$ \textit{open-ended} questions that require deeper engagement, such as \textit{hypothesis testing, causal analysis, scenario completion}, and \textit{role-playing}. These 11 different open-ended question types, as detailed in Table~\ref{tab:question_types}, assess LVLMs' depth of cultural comprehension, requiring models not merely to recall cultural facts but also to generate nuanced explanations, reason through scenarios, and contextualize cultural elements within coherent, culturally informed narratives.

\textbf{\pearlt.} A streamlined subset comprising $893$ Q\&A pairs (591 closed-form and 302 open-ended questions), \pearlt is randomly sampled from Pearl while maintaining a balance of question types and countries. It is designed to facilitate efficient evaluation of proprietary models and minimize costly API usage.

\section{Experimental Setup}
\label{sec:experiments}

\subsection{Models}
For evaluation of~\ourdataset, we use both open and proprietary models (accessible via API calls). We use open models ranging in size between $3$ billion to $72$ billion parameters. To ensure a level playing field, all open models received the same input, sampling configurations, and temperature during the generation process. The specific models we evaluate are \texttt{Qwen$2.5$-VL}~\cite{bai2025qwen2}, \texttt{Aya-Vision}~\cite{dash2025aya}, \texttt{Gemma$3$}~\cite{team2025gemma}, \texttt{Gemeni$2.5$ Pro}~\cite{google2025gemini2.5pro}, \texttt{Claude Sonnet 4}~\cite{anthropic2025system}, and \texttt{o$3$ (OpenAI})~\cite{openai_o3_system_card_2025}. To promote reproducibility and future studies, we are making the complete inference logs for each evaluated model publicly available.

\subsection{Evaluation Protocol}

\subsubsection{LVLM as Judge}
We employ an automatic evaluation method using an \textit{LVLM-as-judge} framework. For this role, we exclusively use \texttt{InternVL3.5-38B}~\cite{wang2025internvl3}, a model distinguished by its advanced reasoning capabilities. This model is part of the state-of-the-art InternVL3.5 family, which has demonstrated top-tier performance among open-source MLLMs across general multimodal, reasoning, text, and agentic tasks. 

\textbf{Metrics.}
Our evaluation protocol adopts two distinct scoring methods, each tailored to a specific question format. For \textbf{\textit{closed-form}} questions (e.g., multiple-choice and True/False), we utilize a \textit{relaxed-match accuracy} (ACC) metric. Here, the judge assesses semantic equivalence between candidate responses and gold-standard answers, permitting synonyms, paraphrases, or minor lexical variations. Each response is assigned a binary correctness score ($1$ for correct, $0$ for incorrect), aggregated into an overall accuracy. For \textbf{\textit{open-ended}} questions—comprising 11 varied types such as cause-and-effect, comparative analysis, and scenario completion—the judge evaluates responses using a comprehensive structured rubric capturing four critical dimensions: \textit{correctness}, \textit{coherence}, \textit{detail}, and \textit{fluency}.\footnote{See Figure~\ref{appdx_fig:prompt_eval} for detailed evaluation prompts defining these criteria explicitly.} Each dimension is scored individually on a scale from $1$ to $5$, with an aggregated, weighted \textbf{\textit{Overall Score}} calculated as follows:
\begin{equation}
\begin{aligned}
\mathit{Overall\ Score} &= 0.4\,\mathit{Correctness}\\
                        &\quad+ 0.2\,\mathit{Coherence}\\
                        &\quad+ 0.2\,\mathit{Detail}\\
                        &\quad+ 0.2\,\mathit{Fluency}
\end{aligned}
\end{equation}

\paragraph{}
\noindent We also follow~\newcite{burda2025culturally} in employing a \textit{Cultural Awareness Score (CAS)}. CAS is a binary metric (0/1) indicating explicitly whether the candidate response mentions culturally-specific elements required by the reference answer, ensuring explicit cultural grounding in the evaluations.
 
\subsubsection{Human Evaluation}
We evaluate using a subset of \pearlt  open-ended dataset (N=70). To obtain human evaluations, we recruited four native Arabic speakers with deep familiarity in the relevant cultural contexts, who independently scored each sample according to the same evaluation rubric applied by our LVLM judges. The user interface utilized by human evaluators can be found in Appendix~\ref{fig:eval_screen}.

To assess the reliability of our human annotations and LVLM judge, we conducted Intraclass Correlation Coefficient (ICC)~\cite{shrout1979intraclass} analysis on the overall evaluation scores. First, we measured inter-annotator agreement among the four human evaluators, which revealed moderate reliability for individual raters ($ICC_{(3,1)} = 0.537, 95\%\,\text{Confidence Interval (CI) } [0.42, 0.65]$) but good reliability when averaged across all annotators ($ICC_{(3,k)} = 0.823, 95\%\,\text{CI}\,[0.74, 0.88]$). Subsequently, we evaluated the agreement between our LVLM judge and the consensus human ratings (averaged across the four annotators) to establish the reliability of our automated evaluation approach. The model demonstrated good agreement with human consensus ($ICC_{(3,1)} = 0.708, 95\%\,\text{CI}\,[0.57, 0.81]$), demonstrating that our LVLM judge can produce evaluations that are reasonably consistent with human expert judgment.

\section{Results and Discussion}
\label{sec:results}

\begin{figure*}[!t]
    \centering
    \includegraphics[width=0.99\linewidth]{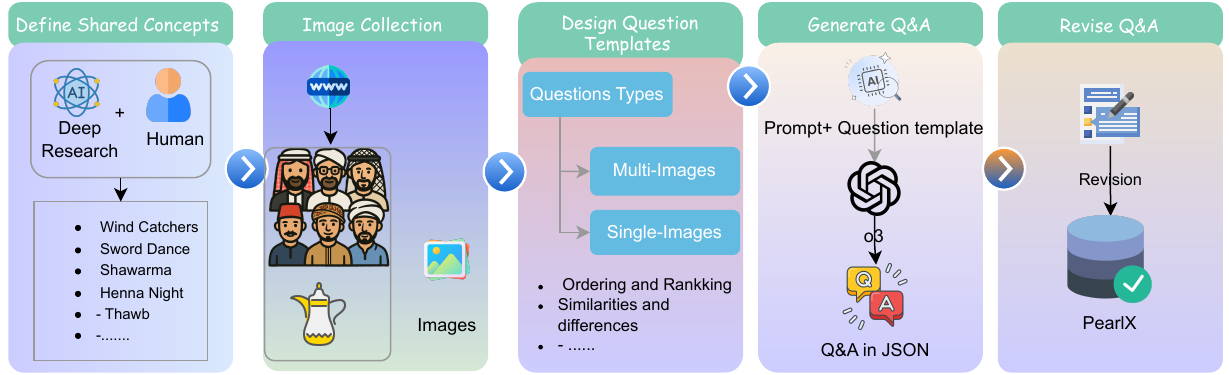}
    \caption{Workflow generation of~\pearlx.}
    \label{fig:shared_workflow}
\end{figure*}

\begin{figure*}[!t]
    \centering
    \includegraphics[width=0.48\linewidth]{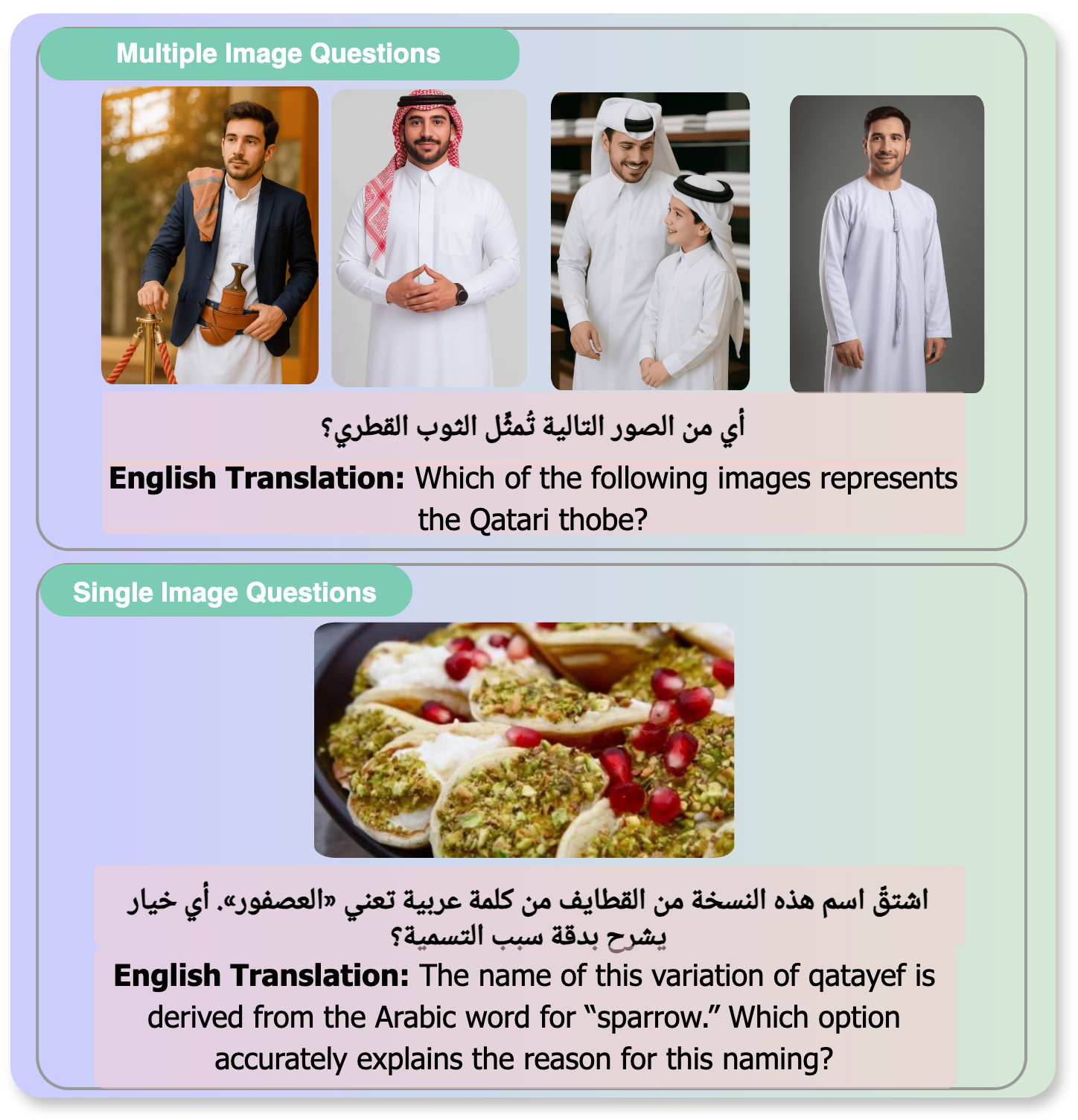}
    \caption{\pearlx exemplars for the Multiple-Image (top) and Single-Image (bottom) question formats. Each prompt is shown in Arabic with its English translation beneath. All depicted faces are synthetic and do not portray real individuals.}
    \label{fig:pearlx_examples}
\end{figure*}

\subsection{Open Models}
\label{sec:open-model-results}

Table~\ref{tab:main_results} presents the performance of nine openly-accessible LVLMs on the \ourdataset benchmark.  
We organize the discussion around two axes: (\textit{i}) parameter scaling within the same family and (\textit{ii}) cultural grounding and closed-form accuracy.

\paragraph{Scaling within the same family.} Within the \texttt{Qwen2.5-VL} line, performance steadily improves as model size grows: the overall score rises from $2.74$ for the 3B-parameter instruct model to $3.16$ for the 7B variant, reaching $3.46$ for the 72B model. A similar upward trajectory is seen in the CAS, which climbs from $33.70$\% to $44.79$\% and then to $49.63$\%. Interestingly, the \texttt{Qwen2.5-VL-32B-Instruct}\footnote{The official release notes stating that Qwen2.5-VL-32B is specifically designed for enhanced reasoning and closer alignment with human preferences; see \url{https://qwenlm.github.io/blog/qwen2.5-vl-32b/}.} reasoning model breaks this trend, outperforming even the 72B variant with an overall score of $3.77$ and a CAS of $67.05$\%. This suggests that architectural or training enhancements in the reasoning variant provide stronger cultural grounding and open-ended response quality than parameter scaling alone.

\paragraph{Cross-family comparison highlights stylistic trade-offs.}
The \texttt{Gemma-3} series delivers balanced results despite lacking dedicated reasoning training.  
Its 12B and 27B variants achieve CAS values around $48-56$~higher than \texttt{Aya-Vision-8/32B}—and competitive Overall scores with other models.  
\texttt{Aya-Vision-32B}, on the other hand, favors fluent, stylistically polished answers (FLU~4.29) but lags in cultural specificity (CAS~$51.2$).  
These contrasts confirm that model design choices (pretraining corpora, vision encoder quality, alignment objectives) influence different quality axes in complementary ways.

\paragraph{Take-away.}
Among the \textit{nine} open models, the clear front-runner is \texttt{Qwen2.5-VL-32B-Instruct}.  
It combines the highest Overall score ($3.77$) with the strongest closed-form accuracy ($79.8$\%) and the best CAS ($67.1$\%) of any system in Table \ref{tab:lite_results}.  
For downstream Arabic-cultural applications where access to proprietary systems is not feasible, \texttt{Qwen2.5-VL-32B} therefore offers the most reliable balance of factual correctness and cultural grounding.  
Looking ahead, further progress is likely to come from reasoning-centered alignment and culturally informed pre-training, rather than from simply adding more parameters.

\begin{table}[t]
  \centering
  \small
  \resizebox{\columnwidth}{!}{%
    \begin{tabular}{@{} l *{7}{c} @{}}
      \toprule
      & \multicolumn{6}{c}{\textbf{Open-Ended}} & \textbf{Closed} \\
      \cmidrule(lr){2-7} \cmidrule(lr){8-8}
      \textbf{Model} & \textbf{COR} & \textbf{COH} & \textbf{DET} & \textbf{FLU} & \textbf{OVR} & \textbf{CAS\%} & \textbf{ACC\%} \\
      \midrule
      Qwen2.5-VL-3B-Instruct & $2.48$ & $3.07$ & $2.05$ & $3.60$ & $2.74$ & $33.70$ & $71.15$ \\
      gemma-3-4b-it          & $2.86$ & $3.43$ & $2.54$ & $4.06$ & $3.15$ & $44.72$ & $69.78$ \\
      \addlinespace[0.1em]\cdashline{1-8}\addlinespace[0.1em]

      Qwen2.5-VL-7B-Instruct & $2.89$ & $3.55$ & $2.45$ & $4.02$ & $3.16$ & $44.79$ & $73.01$ \\
      aya-vision-8b          & $3.17$ & $3.77$ & $2.62$ & $4.21$ & $3.39$ & $45.34$ & $69.89$ \\
      gemma-3-12b-it         & $2.98$ & $3.54$ & $2.54$ & $4.10$ & $3.23$ & $48.13$ & $75.83$ \\
      \addlinespace[0.1em]\cdashline{1-8}\addlinespace[0.1em]

      gemma-3-27b-it         & $3.26$ & $3.76$ & $2.82$ & $4.23$ & $3.47$ & $55.75$ & $79.86$ \\
      aya-vision-32b         & $3.36$ & $3.91$ & $2.75$ & $4.29$ & $3.53$ & $51.19$ & $79.12$ \\
      Qwen2.5-VL-32B-Instruct~\dsdiamond{ftcolor} & $3.61$ & $3.99$ & $3.30$ & $4.32$ & $3.77$ & $67.05$ & $79.78$ \\
      \addlinespace[0.1em]\cdashline{1-8}\addlinespace[0.1em]

      Qwen2.5-VL-72B-Instruct & $3.27$ & $3.82$ & $2.71$ & $4.21$ & $3.46$ & $49.63$ & $79.55$ \\
      \bottomrule
    \end{tabular}%
  }
  \caption{Performance on the \ourdataset benchmark for open models. Open-ended metrics (COR, COH, DET, FLU, OVR) are averaged on a 1--5 scale, 
while CAS\% and ACC\% are reported as percentages. A \textcolor{ftcolor}{$\diamond$} marks models that use explicit reasoning techniques in this paper}
  \label{tab:main_results}
\end{table}

\subsection{Proprietary Models}
\begin{table}[t]
  \centering
  \small
  \resizebox{\columnwidth}{!}{%
    \begin{tabular}{@{} l *{7}{c} @{}}
      \toprule
      & \multicolumn{6}{c}{\textbf{Open-Ended}} & \textbf{Closed} \\
      \cmidrule(lr){2-7} \cmidrule(lr){8-8}
      \textbf{Model} & \textbf{COR} & \textbf{COH} & \textbf{DET} & \textbf{FLU} & \textbf{OVR} & \textbf{CAS\%} & \textbf{ACC\%} \\
      \midrule
      Qwen2.5-VL-3B-Instruct      & $2.59$ & $3.12$ & $2.15$ & $3.68$ & $2.83$ & $37.09$ & $73.10$ \\
      gemma-3-4b-it               & $2.97$ & $3.56$ & $2.62$ & $4.12$ & $3.25$ & $47.68$ & $70.73$ \\
      \addlinespace[0.1em]\cdashline{1-8}\addlinespace[0.1em]
      Qwen2.5-VL-7B-Instruct      & $3.02$ & $3.66$ & $2.55$ & $4.11$ & $3.27$ & $47.68$ & $73.77$ \\
      aya-vision-8b               & $3.23$ & $3.85$ & $2.67$ & $4.21$ & $3.44$ & $46.69$ & $70.56$ \\
      gemma-3-12b-it              & $3.10$ & $3.63$ & $2.54$ & $4.14$ & $3.30$ & $52.65$ & $76.82$ \\
      \addlinespace[0.1em]\cdashline{1-8}\addlinespace[0.1em]
      gemma-3-27b-it              & $3.38$ & $3.81$ & $2.95$ & $4.26$ & $3.55$ & $60.93$ & $80.88$ \\
      aya-vision-32b              & $3.37$ & $3.92$ & $2.76$ & $4.31$ & $3.55$ & $51.66$ & $75.63$ \\
      Qwen2.5-VL-32B-Instruct~\dsdiamond{ftcolor}      & $3.69$ & $4.09$ & $3.39$ & $4.42$ & $3.85$ & $66.56$ & $80.03$ \\
      \addlinespace[0.1em]\cdashline{1-8}\addlinespace[0.1em]
      Qwen2.5-VL-72B-Instruct      & $3.36$ & $3.91$ & $2.76$ & $4.25$ & $3.53$ & $55.63$ & $79.36$ \\
      \hline
      claude-sonnet-4-20250514~\dsdiamond{ftcolor}     & $3.77$ & $4.03$ & $3.71$ & $4.43$ & $3.94$ & $76.49$ & $79.53$ \\
      gemini-2.5-pro~\dsdiamond{ftcolor}        & $4.36$ & $4.48$ & $4.45$ & $4.77$ & $4.48$ & $83.11$ & $89.00$ \\
      o3-2025-04-16~\dsdiamond{ftcolor}       & $4.39$ & $4.44$ & $4.52$ & $4.71$ & $4.49$ & $87.09$ & $86.97$ \\
      \bottomrule
    \end{tabular}%
  }
  \caption{Results for the \pearlt subset. A \textcolor{ftcolor}{$\diamond$} marks models that use explicit reasoning techniques in this paper.}
  \label{tab:lite_results}
\end{table}

Results on the \pearlt benchmark (Table~\ref{tab:lite_results}) clearly demonstrate the superiority of proprietary models (\texttt{Gemini Pro} and \texttt{OpenAI o3}) across all evaluation dimensions. In particular, \texttt{OpenAI o3} achieves the highest overall score of $4.49$ and the highest CAS of $87\%$. Furthermore, proprietary models consistently outperform open models on closed-form question accuracy, with  \texttt{Gemini2.5 Pro} achieving $89\%$, surpassing the best-performing open model (\texttt{gemma-3-27b-it}) at $81\%$.

\section{\pearlx Benchmark}
\label{sec:pearlx}

\begin{table}[t]
  \centering
  \small
  \resizebox{0.28\textwidth}{!}{%
    \begin{tabular}{@{} l c @{}}
      \toprule
      \textbf{Model} & \textbf{Accuracy \%} \\
      \midrule
      Qwen2.5-VL-3B-Instruct      & $59.67$ \\
      gemma-3-4b-it               & $64.58$ \\
      \addlinespace[0.1em]\cdashline{1-2}\addlinespace[0.1em]

      Qwen2.5-VL-7B-Instruct      & $61.31$ \\
      aya-vision-8b               & $64.03$ \\
      gemma-3-12b-it              & $69.21$ \\
      \addlinespace[0.1em]\cdashline{1-2}\addlinespace[0.1em]

      gemma-3-27b-it              & $69.21$ \\
      aya-vision-32b              & $71.66$ \\
      Qwen2.5-VL-32B-Instruct~\dsdiamond{ftcolor} & $71.66$ \\
      \addlinespace[0.1em]\cdashline{1-2}\addlinespace[0.1em]

      Qwen2.5-VL-72B-Instruct      & $73.57$ \\
      \hline
      gemini-2.5-pro-preview-05-06~\dsdiamond{ftcolor}       & $77.93$ \\
      o3-2025-04-16~\dsdiamond{ftcolor}        & $78.75$ \\
      \bottomrule
    \end{tabular}%
  }
  \caption{Accuracy on the \pearlx shared-concepts benchmark.}
  \label{tab:pearl_shared_results}
\end{table}

While working on~\ourdataset, we observed existence of cultural element or practices prevalent across multiple regions or countries that, despite a fundamental similarity, exhibit subtle local variations in appearance, preparation, usage, or style. For instance, the traditional headband known as the \textit{agal} "$\RL{العقال}$", widely worn in countries such as the UAE, Saudi Arabia, and other Gulf nations, displays regional differences in shape, color, and style. Motivated by this cultural insight and supported iteratively by the \texttt{ChatGPT-o$3$} model,\footnote{We used the model to generate the initial questions.} we manually identified $61$ such culturally shared concepts, each observed across at least two\footnote{For example, the traditional dish \textit{kabsa} "$\RL{كبسة}$" appears widely in seven Arab countries, with notable regional variations.} Arab countries.\footnote{Initially, we compiled a preliminary list of approximately $100$ potential shared concepts. After thorough manual filtering to exclude irrelevant or inaccurately represented concepts, we finalized a refined set of $61$ authentic cultural concepts.} A complete list of the identified shared concepts is provided in Appendix~\ref{tab:shared_concepts_elemnts}. We then manually located images from publicly accessible web resources. On average, we gathered approximately three representative images per concept, for a total of $347$ images, ensuring a rich visual depiction of cultural diversity. We then developed \textit{MCQ} and \textit{True/False} questions exploiting the collected images. We include two categories of questions, differing based on whether we feed the model a single image or multiple images.  \textit{Single-image} questions focus on aspects specific to one country through an individual image, whereas \textit{multiple-image} questions target variations among countries regarding the particular shared concept depicted across several images. 
The next step was to generate questions based on manually developed templates that we provide to \texttt{ChatGPT-o$3$} along with the images and the name of each shared concept, enabling it to generate relevant questions. The prompt we developed to generate the questions is shown in Appendix ~\ref{appdx_fig:prompt_new}. In total, we produce $367$ questions, split into $177$ single-image and $190$ multiple-image questions. Finally, we conduct a thorough human review of all generated questions, ensuring that the questions and answers are accurate, meaningful, error-free, and diverse. Figure~\ref{fig:shared_workflow} demonstrates the workflow used to develop \pearlx benchmark. Figure~\ref{fig:pearlx_examples} shows examples of the shared concepts for both single and multiple image questions. Additionally, Appendix ~\ref{appdx_tab:templete_examples} presents sample questions based on one-image and multi-image templates used for generating shared concept questions.

\subsection{Evaluation on \pearlx}
As shown in Table~\ref{tab:pearl_shared_results}, proprietary models (\texttt{Gemini 2.5 Pro} and \texttt{o3}) demonstrate superior performance, with \texttt{o3}-2025-04-16 achieving the highest accuracy ($78.75$\%). Among the open-source models, \texttt{Qwen2.5-VL-72B-Instruct} performs best, reaching an accuracy of $73.57$\%. Notably, reasoning-centric models generally outperform their counterparts, emphasizing the critical role of explicit reasoning alignment in culturally sensitive contexts.
\section{Conclusion}
\label{sec:conclusion}
In this paper, we presented \ourdataset, an Arabic multimodal instruction dataset and benchmarking suite tailored to enhance the cultural understanding capabilities of large vision-language models. \ourdataset addresses critical gaps in existing resources by encompassing diverse culturally-authentic materials across ten domains from the Arab world. Our rigorous annotation and agentic workflows, combined with the expertise of \annotators local annotators, ensure high-quality, culturally-relevant content. Comprehensive evaluations confirm the superior performance of models explicitly optimized for reasoning tasks over parameter scaling, underscoring the importance of culturally-aware alignment methods. The specialized \pearlx benchmark further allows nuanced assessment of cross-country cultural variations, setting the stage for more culturally sensitive model development.

\section*{Limitations}
Despite its comprehensive scope, the \ourdataset dataset has a number of limitations. First, the dataset predominantly relies on publicly available resources like Wikipedia, potentially introducing biases towards topics and perspectives that are well-documented online. Second, while we involved annotators from diverse Arab countries, the coverage does not equally represent all regions. Additionally, although we employed rigorous human-in-the-loop annotation processes, subjective cultural interpretations may still influence data annotation consistency. Lastly, due to the focus on cultural specificity, generalizability to other non-Arabic cultures or languages may be limited, requiring additional datasets and evaluations tailored to different cultural contexts.

\section*{Ethics Statement}
In developing \ourdataset, we emphasized cultural sensitivity, inclusivity, and ethical responsibility. All annotations were created by informed participants, each of whom is acknowledged and credited as a contributor. We adhered strictly to publicly available and reputable sources, refraining from using any private or sensitive data. Clear guidelines were provided to respect local norms, maintain data privacy, and secure participant consent. All images utilized in this dataset are sourced from Wikipedia under Creative Commons licenses, and any images originating outside Wikipedia have been masked or regenerated with alternative identities to ensure privacy and ethical compliance.

Although \ourdataset aims to mitigate biases in Arabic LVLMs, unintentional cultural bias may still occur—particularly in regions lacking direct local representation. We encourage ongoing community involvement to address these gaps, ensuring continual refinement and improvement of the dataset.

\paragraph{Reproducibility.} Our test data, prompts, and code necessary to produce all results reported in this work are publicly available at \url{https://github.com/UBC-NLP/pearl}.

\section*{Acknowledgments}\label{sec:acknow}
We acknowledge support from Canada Research Chairs (CRC), the Natural Sciences and Engineering Research Council of Canada (NSERC; RGPIN-2018-04267), the Social Sciences and Humanities Research Council of Canada (SSHRC; 895-2020-1004; 895-2021-1008), Canadian Foundation for Innovation (CFI; 37771), Digital Research Alliance of Canada,\footnote{\href{https://alliancecan.ca}{https://alliancecan.ca}} and UBC Advanced Research Computing-Sockeye.\footnote{\href{https://arc.ubc.ca/ubc-arc-sockeye}{https://arc.ubc.ca/ubc-arc-sockeye}} We also acknowledge the valuable contributions of Razan Khassib, Lina Hamad, Fatimah Alshamari, Cheikh Malainine, Doaa Qawasmeh, Tfeil Moilid, Sara Shatnawi, and Ahmed Aboeitta.

\bibliography{custom}

\begin{thebibliography}{56}
\providecommand{\natexlab}[1]{#1}

\bibitem[{Aloui et~al.(2024)Aloui, Chouikhi, Chaabane, Kchaou, and Dhaouadi}]{aloui2024101}
Manel Aloui, Hasna Chouikhi, Ghaith Chaabane, Haithem Kchaou, and Chehir Dhaouadi. 2024.
\newblock \href {https://arxiv.org/abs/2405.01590} {101 billion arabic words dataset}.
\newblock \emph{arXiv preprint arXiv:2405.01590}.

\bibitem[{Alwajih et~al.(2024)Alwajih, Nagoudi, Bhatia, Mohamed, and Abdul-Mageed}]{alwajih2024peacock}
Fakhraddin Alwajih, El~Moatez~Billah Nagoudi, Gagan Bhatia, Abdelrahman Mohamed, and Muhammad Abdul-Mageed. 2024.
\newblock \href {https://arxiv.org/abs/2403.01031} {Peacock: A family of arabic multimodal large language models and benchmarks}.
\newblock \emph{arXiv preprint arXiv:2403.01031}.

\bibitem[{Alyafeai et~al.(2024)Alyafeai, Almubarak, Ashraf, Alnuhait, Alshahrani, Abdulrahman, Ahmed, Gawah, Saleh, Ghaleb et~al.}]{alyafeai2024cidar}
Zaid Alyafeai, Khalid Almubarak, Ahmed Ashraf, Deema Alnuhait, Saied Alshahrani, Gubran~AQ Abdulrahman, Gamil Ahmed, Qais Gawah, Zead Saleh, Mustafa Ghaleb, and 1 others. 2024.
\newblock \href {https://arxiv.org/abs/2402.03177} {Cidar: Culturally relevant instruction dataset for arabic}.
\newblock \emph{arXiv preprint arXiv:2402.03177}.

\bibitem[{Ananthram et~al.(2024)Ananthram, Stengel-Eskin, Vondrick, Bansal, and McKeown}]{ananthram2024see}
Amith Ananthram, Elias Stengel-Eskin, Carl Vondrick, Mohit Bansal, and Kathleen McKeown. 2024.
\newblock \href {https://arxiv.org/abs/2406.11665} {See it from my perspective: Diagnosing the western cultural bias of large vision-language models in image understanding}.
\newblock \emph{arXiv preprint arXiv:2406.11665}.

\bibitem[{Anthropic(2025)}]{anthropic2025system}
Anthropic. 2025.
\newblock \href {https://www-cdn.anthropic.com/4263b940cabb546aa0e3283f35b686f4f3b2ff47.pdf} {System card: Claude opus 4 \& claude sonnet 4}.
\newblock System card, Anthropic.

\bibitem[{Baek et~al.(2024)Baek, Park, Kim, Heo, Chang, and Choo}]{baek2024evaluating}
Yujin Baek, ChaeHun Park, Jaeseok Kim, Yu-Jung Heo, Du-Seong Chang, and Jaegul Choo. 2024.
\newblock \href {https://arxiv.org/abs/2406.16469} {Evaluating visual and cultural interpretation: The k-viscuit benchmark with human-vlm collaboration}.
\newblock \emph{arXiv preprint arXiv:2406.16469}.

\bibitem[{Bai et~al.(2025)Bai, Chen, Liu, Wang, Ge, Song, Dang, Wang, Wang, Tang et~al.}]{bai2025qwen2}
Shuai Bai, Keqin Chen, Xuejing Liu, Jialin Wang, Wenbin Ge, Sibo Song, Kai Dang, Peng Wang, Shijie Wang, Jun Tang, and 1 others. 2025.
\newblock \href {https://arxiv.org/abs/2502.13923} {Qwen2. 5-vl technical report}.
\newblock \emph{arXiv preprint arXiv:2502.13923}.

\bibitem[{Becattini et~al.(2023)Becattini, Bongini, Bulla, Bimbo, Marinucci, Mongiov{\`\i}, and Presutti}]{becattini2023viscounth}
Federico Becattini, Pietro Bongini, Luana Bulla, Alberto~Del Bimbo, Ludovica Marinucci, Misael Mongiov{\`\i}, and Valentina Presutti. 2023.
\newblock \href {https://dl.acm.org/doi/10.1145/3580570} {Viscounth: a large-scale multilingual visual question answering dataset for cultural heritage}.
\newblock \emph{ACM Transactions on Multimedia Computing, Communications and Applications}, 19(6):1--20.

\bibitem[{Bhatia et~al.(2024)Bhatia, Ravi, Chinchure, Hwang, and Shwartz}]{bhatia2024local}
Mehar Bhatia, Sahithya Ravi, Aditya Chinchure, Eunjeong Hwang, and Vered Shwartz. 2024.
\newblock \href {https://arxiv.org/abs/2407.00263} {From local concepts to universals: Evaluating the multicultural understanding of vision-language models}.
\newblock \emph{arXiv preprint arXiv:2407.00263}.

\bibitem[{Burda-Lassen et~al.(2024)Burda-Lassen, Chadha, Goswami, and Jain}]{burda2024culturally}
Olena Burda-Lassen, Aman Chadha, Shashank Goswami, and Vinija Jain. 2024.
\newblock \href {https://arxiv.org/abs/2405.17475} {How culturally aware are vision-language models?}
\newblock \emph{arXiv preprint arXiv:2405.17475}.

\bibitem[{Burda-Lassen et~al.(2025)Burda-Lassen, Chadha, Goswami, and Jain}]{burda2025culturally}
Olena Burda-Lassen, Aman Chadha, Shashank Goswami, and Vinija Jain. 2025.
\newblock \href {https://arxiv.org/pdf/2405.17475} {How culturally aware are vision-language models?}
\newblock In \emph{2025 IEEE 6th International Conference on Image Processing, Applications and Systems (IPAS)}, pages 1--6. IEEE.

\bibitem[{Changpinyo et~al.(2022)Changpinyo, Xue, Yarom, Thapliyal, Szpektor, Amelot, Chen, and Soricut}]{changpinyo2022maxm}
Soravit Changpinyo, Linting Xue, Michal Yarom, Ashish~V Thapliyal, Idan Szpektor, Julien Amelot, Xi~Chen, and Radu Soricut. 2022.
\newblock \href {https://arxiv.org/abs/2209.05401} {Maxm: Towards multilingual visual question answering}.
\newblock \emph{arXiv preprint arXiv:2209.05401}.

\bibitem[{Chen et~al.(2024)Chen, Xu, Kirmani, Ichter, Sadigh, Guibas, and Xia}]{Chen_2024_CVPR}
Boyuan Chen, Zhuo Xu, Sean Kirmani, Brain Ichter, Dorsa Sadigh, Leonidas Guibas, and Fei Xia. 2024.
\newblock \href {https://openaccess.thecvf.com/content/CVPR2024/papers/Chen_SpatialVLM_Endowing_Vision-Language_Models_with_Spatial_Reasoning_Capabilities_CVPR_2024_paper.pdf} {Spatialvlm: Endowing vision-language models with spatial reasoning capabilities}.
\newblock In \emph{Proceedings of the IEEE/CVF Conference on Computer Vision and Pattern Recognition (CVPR)}, pages 14455--14465.

\bibitem[{Das et~al.(2024)Das, Hristov, Li, Dimitrov, Koychev, and Nakov}]{das2024exams}
Rocktim~Jyoti Das, Simeon~Emilov Hristov, Haonan Li, Dimitar~Iliyanov Dimitrov, Ivan Koychev, and Preslav Nakov. 2024.
\newblock \href {https://arxiv.org/abs/2403.10378} {Exams-v: A multi-discipline multilingual multimodal exam benchmark for evaluating vision language models}.
\newblock \emph{arXiv preprint arXiv:2403.10378}.

\bibitem[{Dash et~al.(2025)Dash, Nan, Dang, Ahmadian, Singh, Smith, Venkitesh, Shmyhlo, Aryabumi, Beller-Morales et~al.}]{dash2025aya}
Saurabh Dash, Yiyang Nan, John Dang, Arash Ahmadian, Shivalika Singh, Madeline Smith, Bharat Venkitesh, Vlad Shmyhlo, Viraat Aryabumi, Walter Beller-Morales, and 1 others. 2025.
\newblock \href {https://arxiv.org/abs/2505.08751} {Aya vision: Advancing the frontier of multilingual multimodality}.
\newblock \emph{arXiv preprint arXiv:2505.08751}.

\bibitem[{DeepMind(2025)}]{google2025gemini2.5pro}
Google DeepMind. 2025.
\newblock \href {https://cloud.google.com/vertex-ai/generative-ai/docs/models/gemini/2-5-pro} {Gemini 2.5 pro}.
\newblock Vertex AI Documentation.
\newblock Accessed May 18, 2025.

\bibitem[{Ghaboura et~al.(2024)Ghaboura, Heakl, Thawakar, Alharthi, Riahi, Saif, Laaksonen, Khan, Khan, and Anwer}]{ghaboura2024camel}
Sara Ghaboura, Ahmed Heakl, Omkar Thawakar, Ali Alharthi, Ines Riahi, Abduljalil Saif, Jorma Laaksonen, Fahad~S. Khan, Salman Khan, and Rao~M. Anwer. 2024.
\newblock \href {https://arxiv.org/abs/2410.18976} {Camel-bench: A comprehensive arabic lmm benchmark}.
\newblock \emph{Preprint}, arXiv:2410.18976.

\bibitem[{Han et~al.(2023)Han, You, Liu, Chen, Zheng, Mrini, Lin, Wang, Zhai, Yuan et~al.}]{han2023infimm}
Xiaotian Han, Quanzeng You, Yongfei Liu, Wentao Chen, Huangjie Zheng, Khalil Mrini, Xudong Lin, Yiqi Wang, Bohan Zhai, Jianbo Yuan, and 1 others. 2023.
\newblock \href {https://arxiv.org/abs/2311.11567} {Infimm-eval: Complex open-ended reasoning evaluation for multi-modal large language models}.
\newblock \emph{arXiv preprint arXiv:2311.11567}.

\bibitem[{Hendrycks et~al.(2020)Hendrycks, Burns, Basart, Zou, Mazeika, Song, and Steinhardt}]{hendrycks2020measuring}
Dan Hendrycks, Collin Burns, Steven Basart, Andy Zou, Mantas Mazeika, Dawn Song, and Jacob Steinhardt. 2020.
\newblock \href {https://arxiv.org/abs/2009.03300} {Measuring massive multitask language understanding}.
\newblock \emph{arXiv preprint arXiv:2009.03300}.

\bibitem[{Huang et~al.(2024)Huang, Yu, Zhu, Sun, Cheng, Dingjie, Chen, Alharthi, An, He et~al.}]{huang2024acegpt}
Huang Huang, Fei Yu, Jianqing Zhu, Xuening Sun, Hao Cheng, Song Dingjie, Zhihong Chen, Mosen Alharthi, Bang An, Juncai He, and 1 others. 2024.
\newblock \href {https://aclanthology.org/2024.naacl-long.453/} {Acegpt, localizing large language models in arabic}.
\newblock In \emph{Proceedings of the 2024 Conference of the North American Chapter of the Association for Computational Linguistics: Human Language Technologies (Volume 1: Long Papers)}, pages 8132--8156.

\bibitem[{Huang et~al.(2023)Huang, Yu, Zhu, Sun, Cheng, Song, Chen, Alharthi, An, Liu et~al.}]{huang2023acegpt}
Huang Huang, Fei Yu, Jianqing Zhu, Xuening Sun, Hao Cheng, Dingjie Song, Zhihong Chen, Abdulmohsen Alharthi, Bang An, Ziche Liu, and 1 others. 2023.
\newblock \href {https://arxiv.org/abs/2309.12053} {Acegpt, localizing large language models in arabic}.
\newblock \emph{arXiv preprint arXiv:2309.12053}.

\bibitem[{Kadaoui et~al.(2025)Kadaoui, Atwany, Al-Ali, Mohamed, Mekky, Tilga, Fedorova, Artemova, Aldarmaki, and Kementchedjhieva}]{kadaoui2025jeem}
Karima Kadaoui, Hanin Atwany, Hamdan Al-Ali, Abdelrahman Mohamed, Ali Mekky, Sergei Tilga, Natalia Fedorova, Ekaterina Artemova, Hanan Aldarmaki, and Yova Kementchedjhieva. 2025.
\newblock \href {https://arxiv.org/abs/2503.21910} {Jeem: Vision-language understanding in four arabic dialects}.
\newblock \emph{arXiv preprint arXiv:2503.21910}.

\bibitem[{Liu et~al.(2025{\natexlab{a}})Liu, Gurevych, and Korhonen}]{liu2025culturally}
Chen~Cecilia Liu, Iryna Gurevych, and Anna Korhonen. 2025{\natexlab{a}}.
\newblock \href {https://direct.mit.edu/tacl/article-pdf/doi/10.1162/tacl_a_00760/2535104/tacl_a_00760.pdf} {Culturally aware and adapted nlp: A taxonomy and a survey of the state of the art}.
\newblock \emph{Transactions of the Association for Computational Linguistics}, 13:652--689.

\bibitem[{Liu et~al.(2021)Liu, Bugliarello, Ponti, Reddy, Collier, and Elliott}]{liu2021visually}
Fangyu Liu, Emanuele Bugliarello, Edoardo~Maria Ponti, Siva Reddy, Nigel Collier, and Desmond Elliott. 2021.
\newblock \href {https://arxiv.org/abs/2109.13238} {Visually grounded reasoning across languages and cultures}.
\newblock \emph{arXiv preprint arXiv:2109.13238}.

\bibitem[{Liu et~al.(2025{\natexlab{b}})Liu, Jin, Li, Wong, Wen, Sun, Chen, Xie, and Wang}]{liu2025culturevlm}
Shudong Liu, Yiqiao Jin, Cheng Li, Derek~F Wong, Qingsong Wen, Lichao Sun, Haipeng Chen, Xing Xie, and Jindong Wang. 2025{\natexlab{b}}.
\newblock \href {https://arxiv.org/abs/2501.01282} {Culturevlm: Characterizing and improving cultural understanding of vision-language models for over 100 countries}.
\newblock \emph{arXiv preprint arXiv:2501.01282}.

\bibitem[{Liu et~al.(2025{\natexlab{c}})Liu, Duan, Zhang, Li, Zhang, Zhao, Yuan, Wang, He, Liu et~al.}]{liu2025mmbench}
Yuan Liu, Haodong Duan, Yuanhan Zhang, Bo~Li, Songyang Zhang, Wangbo Zhao, Yike Yuan, Jiaqi Wang, Conghui He, Ziwei Liu, and 1 others. 2025{\natexlab{c}}.
\newblock \href {https://doi.org/10.1007/978-3-031-73002-6_13} {Mmbench: Is your multi-modal model an all-around player?}
\newblock In \emph{European conference on computer vision}, pages 216--233. Springer.

\bibitem[{Ma et~al.(2023)Ma, Pan, Wu, Cheng, Zhang, Huang, and Chen}]{ma2023food}
Zheng Ma, Mianzhi Pan, Wenhan Wu, Kanzhi Cheng, Jianbing Zhang, Shujian Huang, and Jiajun Chen. 2023.
\newblock \href {https://dl.acm.org/doi/10.1145/3581783.3612715} {Food-500 cap: A fine-grained food caption benchmark for evaluating vision-language models}.
\newblock In \emph{Proceedings of the 31st ACM International Conference on Multimedia}, pages 5674--5685.

\bibitem[{Nayak et~al.(2024)Nayak, Jain, Awal, Reddy, Van~Steenkiste, Hendricks, Agrawal et~al.}]{nayak2024benchmarking}
Shravan Nayak, Kanishk Jain, Rabiul Awal, Siva Reddy, Sjoerd Van~Steenkiste, Lisa~Anne Hendricks, Aishwarya Agrawal, and 1 others. 2024.
\newblock \href {https://arxiv.org/abs/2407.10920} {Benchmarking vision language models for cultural understanding}.
\newblock \emph{arXiv preprint arXiv:2407.10920}.

\bibitem[{Nguyen et~al.(2024)Nguyen, Tran, Pham, Nguyen, Nguyen, Nguyen, and Nguyen}]{vannguyen2024ViTextVQA}
Quan~Van Nguyen, Dan~Quang Tran, Huy~Quang Pham, Thang Kien-Bao Nguyen, Nghia~Hieu Nguyen, Kiet~Van Nguyen, and Ngan Luu-Thuy Nguyen. 2024.
\newblock \href {https://arxiv.org/abs/2404.10652} {Vitextvqa: A large-scale visual question answering dataset for evaluating vietnamese text comprehension in images}.
\newblock \emph{Preprint}, arXiv:2404.10652.

\bibitem[{Onohara et~al.(2024)Onohara, Miyai, Imajuku, Egashira, Baek, Yue, Neubig, and Aizawa}]{onohara2024jmmmu}
Shota Onohara, Atsuyuki Miyai, Yuki Imajuku, Kazuki Egashira, Jeonghun Baek, Xiang Yue, Graham Neubig, and Kiyoharu Aizawa. 2024.
\newblock \href {https://arxiv.org/abs/2410.17250} {Jmmmu: A japanese massive multi-discipline multimodal understanding benchmark for culture-aware evaluation}.
\newblock \emph{Preprint}, arXiv:2410.17250.

\bibitem[{OpenAI(2025)}]{openai_o3_system_card_2025}
OpenAI. 2025.
\newblock \href {https://cdn.openai.com/pdf/2221c875-02dc-4789-800b-e7758f3722c1/o3-and-o4-mini-system-card.pdf} {{OpenAI o3 and o4-mini System Card}}.
\newblock Technical report, OpenAI.
\newblock Version 2 of the Preparedness Framework.

\bibitem[{Parida et~al.(2023)Parida, Abdulmumin, Muhammad, Bose, Kohli, Ahmad, Kotwal, Deb~Sarkar, Bojar, and Kakudi}]{parida-etal-2023-havqa}
Shantipriya Parida, Idris Abdulmumin, Shamsuddeen~Hassan Muhammad, Aneesh Bose, Guneet~Singh Kohli, Ibrahim~Said Ahmad, Ketan Kotwal, Sayan Deb~Sarkar, Ond{\v{r}}ej Bojar, and Habeebah Kakudi. 2023.
\newblock \href {https://doi.org/10.18653/v1/2023.findings-acl.646} {{H}a{VQA}: A dataset for visual question answering and multimodal research in {H}ausa language}.
\newblock In \emph{Findings of the Association for Computational Linguistics: ACL 2023}, pages 10162--10183, Toronto, Canada. Association for Computational Linguistics.

\bibitem[{Pawar et~al.(2024)Pawar, Park, Jin, Arora, Myung, Yadav, Haznitrama, Song, Oh, and Augenstein}]{pawar2024survey}
Siddhesh Pawar, Junyeong Park, Jiho Jin, Arnav Arora, Junho Myung, Srishti Yadav, Faiz~Ghifari Haznitrama, Inhwa Song, Alice Oh, and Isabelle Augenstein. 2024.
\newblock \href {https://arxiv.org/abs/2411.00860} {Survey of cultural awareness in language models: Text and beyond}.
\newblock \emph{arXiv preprint arXiv:2411.00860}.

\bibitem[{Pawar et~al.(2025)Pawar, Park, Jin, Arora, Myung, Yadav, Haznitrama, Song, Oh, and Augenstein}]{pawar2025survey}
Siddhesh Pawar, Junyeong Park, Jiho Jin, Arnav Arora, Junho Myung, Srishti Yadav, Faiz~Ghifari Haznitrama, Inhwa Song, Alice Oh, and Isabelle Augenstein. 2025.
\newblock \href {https://direct.mit.edu/coli/article-pdf/doi/10.1162/COLI.a.14/2523159/coli.a.14.pdf} {Survey of cultural awareness in language models: Text and beyond}.
\newblock \emph{Computational Linguistics}, pages 1--96.

\bibitem[{Pfeiffer et~al.(2021)Pfeiffer, Geigle, Kamath, Steitz, Roth, Vuli{\'c}, and Gurevych}]{pfeiffer2021xgqa}
Jonas Pfeiffer, Gregor Geigle, Aishwarya Kamath, Jan-Martin~O Steitz, Stefan Roth, Ivan Vuli{\'c}, and Iryna Gurevych. 2021.
\newblock \href {https://arxiv.org/abs/2109.06082} {xgqa: Cross-lingual visual question answering}.
\newblock \emph{arXiv preprint arXiv:2109.06082}.

\bibitem[{Romero et~al.(2024)Romero, Lyu, Wibowo, Lynn, Hamed, Kishore, Mandal, Dragonetti, Abzaliev, Tonja et~al.}]{romero2024cvqa}
David Romero, Chenyang Lyu, Haryo~Akbarianto Wibowo, Teresa Lynn, Injy Hamed, Aditya~Nanda Kishore, Aishik Mandal, Alina Dragonetti, Artem Abzaliev, Atnafu~Lambebo Tonja, and 1 others. 2024.
\newblock \href {https://arxiv.org/abs/2406.05967} {Cvqa: Culturally-diverse multilingual visual question answering benchmark}.
\newblock \emph{arXiv preprint arXiv:2406.05967}.

\bibitem[{Schneider et~al.(2025)Schneider, Holtermann, Biemann, and Lauscher}]{schneider2025gimmick}
Florian Schneider, Carolin Holtermann, Chris Biemann, and Anne Lauscher. 2025.
\newblock \href {https://arxiv.org/abs/2502.13766} {Gimmick--globally inclusive multimodal multitask cultural knowledge benchmarking}.
\newblock \emph{arXiv preprint arXiv:2502.13766}.

\bibitem[{Schneider and Sitaram(2024)}]{schneider2024m5}
Florian Schneider and Sunayana Sitaram. 2024.
\newblock \href {https://arxiv.org/abs/2407.03791} {M5--a diverse benchmark to assess the performance of large multimodal models across multilingual and multicultural vision-language tasks}.
\newblock \emph{arXiv preprint arXiv:2407.03791}.

\bibitem[{Sengupta et~al.(2023)Sengupta, Sahu, Jia, Katipomu, Li, Koto, Afzal, Kamboj, Pandit, Pal et~al.}]{sengupta2023jais}
Neha Sengupta, Sunil~Kumar Sahu, Bokang Jia, Satheesh Katipomu, Haonan Li, Fajri Koto, Osama~Mohammed Afzal, Samta Kamboj, Onkar Pandit, Rahul Pal, and 1 others. 2023.
\newblock \href {https://arxiv.org/abs/2308.16149} {Jais and jais-chat: Arabic-centric foundation and instruction-tuned open generative large language models}.
\newblock \emph{arXiv preprint arXiv:2308.16149}.

\bibitem[{Shao et~al.(2024)Shao, Qian, Xiao, Song, Zong, Wang, Liu, and Li}]{NEURIPS2024_0ff38d72}
Hao Shao, Shengju Qian, Han Xiao, Guanglu Song, Zhuofan Zong, Letian Wang, Yu~Liu, and Hongsheng Li. 2024.
\newblock \href {https://proceedings.neurips.cc/paper_files/paper/2024/file/0ff38d72a2e0aa6dbe42de83a17b2223-Paper-Datasets_and_Benchmarks_Track.pdf} {Visual cot: Advancing multi-modal language models with a comprehensive dataset and benchmark for chain-of-thought reasoning}.
\newblock In \emph{Advances in Neural Information Processing Systems}, volume~37, pages 8612--8642. Curran Associates, Inc.

\bibitem[{Shrout and Fleiss(1979)}]{shrout1979intraclass}
Patrick~E Shrout and Joseph~L Fleiss. 1979.
\newblock Intraclass correlations: uses in assessing rater reliability.
\newblock \emph{Psychological bulletin}, 86(2):420.

\bibitem[{Singh et~al.(2024)Singh, Romanou, Fourrier, Adelani, Ngui, Vila-Suero, Limkonchotiwat, Marchisio, Leong, Susanto et~al.}]{singh2024global}
Shivalika Singh, Angelika Romanou, Cl{\'e}mentine Fourrier, David~I Adelani, Jian~Gang Ngui, Daniel Vila-Suero, Peerat Limkonchotiwat, Kelly Marchisio, Wei~Qi Leong, Yosephine Susanto, and 1 others. 2024.
\newblock \href {https://arxiv.org/abs/2412.03304} {Global mmlu: Understanding and addressing cultural and linguistic biases in multilingual evaluation}.
\newblock \emph{arXiv preprint arXiv:2412.03304}.

\bibitem[{Song et~al.(2025)Song, Wu, Zhu, Zhang, and Chen}]{song2025}
Xiujie Song, Mengyue Wu, Kenny~Q. Zhu, Chunhao Zhang, and Yanyi Chen. 2025.
\newblock \href {https://arxiv.org/abs/2402.18409} {A cognitive evaluation benchmark of image reasoning and description for large vision-language models}.
\newblock \emph{Preprint}, arXiv:2402.18409.

\bibitem[{Team et~al.(2025)Team, Kamath, Ferret, Pathak, Vieillard, Merhej, Perrin, Matejovicova, Ram{\'e}, Rivi{\`e}re et~al.}]{team2025gemma}
Gemma Team, Aishwarya Kamath, Johan Ferret, Shreya Pathak, Nino Vieillard, Ramona Merhej, Sarah Perrin, Tatiana Matejovicova, Alexandre Ram{\'e}, Morgane Rivi{\`e}re, and 1 others. 2025.
\newblock \href {https://arxiv.org/abs/2503.19786} {Gemma 3 technical report}.
\newblock \emph{arXiv preprint arXiv:2503.19786}.

\bibitem[{Tkachenko et~al.(2020)Tkachenko, Malyuk, Holmanyuk, and Liubimov}]{tkachenko2020label}
Maxim Tkachenko, Mikhail Malyuk, Andrey Holmanyuk, and Nikolai Liubimov. 2020.
\newblock \href {https://github.com/heartexlabs/label-studio} {Label studio: Data labeling software}.
\newblock \emph{Open source software available from https://github. com/heartexlabs/label-studio}, 2022.

\bibitem[{Vayani et~al.(2024)Vayani, Dissanayake, Watawana, Ahsan, Sasikumar, Thawakar, Ademtew, Hmaiti, Kumar, Kuckreja et~al.}]{vayani2024all}
Ashmal Vayani, Dinura Dissanayake, Hasindri Watawana, Noor Ahsan, Nevasini Sasikumar, Omkar Thawakar, Henok~Biadglign Ademtew, Yahya Hmaiti, Amandeep Kumar, Kartik Kuckreja, and 1 others. 2024.
\newblock \href {https://arxiv.org/abs/2411.16508} {All languages matter: Evaluating lmms on culturally diverse 100 languages}.
\newblock \emph{arXiv preprint arXiv:2411.16508}.

\bibitem[{Wang et~al.(2024{\natexlab{a}})Wang, Xu, Xie, Wang, Li, Xie, Zhang, Xiong, and Chen}]{wang2024m4u}
Hongyu Wang, Jiayu Xu, Senwei Xie, Ruiping Wang, Jialin Li, Zhaojie Xie, Bin Zhang, Chuyan Xiong, and Xilin Chen. 2024{\natexlab{a}}.
\newblock \href {https://arxiv.org/abs/2405.15638} {M4u: Evaluating multilingual understanding and reasoning for large multimodal models}.
\newblock \emph{arXiv preprint arXiv:2405.15638}.

\bibitem[{Wang et~al.(2024{\natexlab{b}})Wang, Ming, Shi, Vineet, Wang, Li, and Joshi}]{wang2024picture}
Jiayu Wang, Yifei Ming, Zhenmei Shi, Vibhav Vineet, Xin Wang, Sharon Li, and Neel Joshi. 2024{\natexlab{b}}.
\newblock \href {https://proceedings.neurips.cc/paper_files/paper/2024/hash/cd3aedacb22c6a5b6b76f5a84df4c7a9-Abstract-Conference.html} {Is a picture worth a thousand words? delving into spatial reasoning for vision language models}.
\newblock \emph{Advances in Neural Information Processing Systems}, 37:75392--75421.

\bibitem[{Wang et~al.(2024{\natexlab{c}})Wang, Cao, Zhang, Yuan, Shan, Chen, and Gao}]{wang2024vlbiasbench}
Sibo Wang, Xiangkui Cao, Jie Zhang, Zheng Yuan, Shiguang Shan, Xilin Chen, and Wen Gao. 2024{\natexlab{c}}.
\newblock \href {https://arxiv.org/abs/2406.14194} {Vlbiasbench: A comprehensive benchmark for evaluating bias in large vision-language model}.
\newblock \emph{Preprint}, arXiv:2406.14194.

\bibitem[{Wang et~al.(2025)Wang, Gao, Gu, Pu, Cui, Wei, Liu, Jing, Ye, Shao et~al.}]{wang2025internvl3}
Weiyun Wang, Zhangwei Gao, Lixin Gu, Hengjun Pu, Long Cui, Xingguang Wei, Zhaoyang Liu, Linglin Jing, Shenglong Ye, Jie Shao, and 1 others. 2025.
\newblock Internvl3. 5: Advancing open-source multimodal models in versatility, reasoning, and efficiency.
\newblock \emph{arXiv preprint arXiv:2508.18265}.

\bibitem[{Wang et~al.(2024{\natexlab{d}})Wang, Chen, Han, Lin, Zhao, Liu, Zhai, Yuan, You, and Yang}]{wang2024exploring}
Yiqi Wang, Wentao Chen, Xiaotian Han, Xudong Lin, Haiteng Zhao, Yongfei Liu, Bohan Zhai, Jianbo Yuan, Quanzeng You, and Hongxia Yang. 2024{\natexlab{d}}.
\newblock \href {https://arxiv.org/abs/2401.06805} {Exploring the reasoning abilities of multimodal large language models (mllms): A comprehensive survey on emerging trends in multimodal reasoning}.
\newblock \emph{arXiv preprint arXiv:2401.06805}.

\bibitem[{Winata et~al.(2024)Winata, Hudi, Irawan, Anugraha, Putri, Wang, Nohejl, Prathama, Ousidhoum, Amriani, Rzayev, Das, Pramodya, Adila, Wilie, Mawalim, Cheng, Abolade, Chersoni, Santus, Ikhwantri, Kuwanto, Zhao, Wibowo, Lovenia, Cruz, Putra, Myung, Susanto, Machin, Zhukova, Anugraha, Adilazuarda, Santosa, Limkonchotiwat, Dabre, Audino, Cahyawijaya, Zhang, Salim, Zhou, Gui, Adelani, Lee, Okada, Purwarianti, Aji, Watanabe, Wijaya, Oh, and Ngo}]{winata2024worldcuisines}
Genta~Indra Winata, Frederikus Hudi, Patrick~Amadeus Irawan, David Anugraha, Rifki~Afina Putri, Yutong Wang, Adam Nohejl, Ubaidillah~Ariq Prathama, Nedjma Ousidhoum, Afifa Amriani, Anar Rzayev, Anirban Das, Ashmari Pramodya, Aulia Adila, Bryan Wilie, Candy~Olivia Mawalim, Ching~Lam Cheng, Daud Abolade, Emmanuele Chersoni, and 32 others. 2024.
\newblock \href {https://arxiv.org/abs/2410.12705} {Worldcuisines: A massive-scale benchmark for multilingual and multicultural visual question answering on global cuisines}.
\newblock \emph{Preprint}, arXiv:2410.12705.

\bibitem[{Ying et~al.(2024)Ying, Meng, Wang, Li, Lin, Yang, Zhang, Zhang, Lin, Liu, Lei, Lu, Chen, Xu, Zhang, Zhang, Gao, Wang, Qiao, Luo, Zhang, and Shao}]{ying2024mmtbench}
Kaining Ying, Fanqing Meng, Jin Wang, Zhiqian Li, Han Lin, Yue Yang, Hao Zhang, Wenbo Zhang, Yuqi Lin, Shuo Liu, Jiayi Lei, Quanfeng Lu, Runjian Chen, Peng Xu, Renrui Zhang, Haozhe Zhang, Peng Gao, Yali Wang, Yu~Qiao, and 3 others. 2024.
\newblock \href {https://arxiv.org/abs/2404.16006} {Mmt-bench: A comprehensive multimodal benchmark for evaluating large vision-language models towards multitask agi}.
\newblock \emph{Preprint}, arXiv:2404.16006.

\bibitem[{Yu et~al.(2023)Yu, Yang, Li, Wang, Lin, Liu, Wang, and Wang}]{yu2023mm}
Weihao Yu, Zhengyuan Yang, Linjie Li, Jianfeng Wang, Kevin Lin, Zicheng Liu, Xinchao Wang, and Lijuan Wang. 2023.
\newblock \href {https://arxiv.org/abs/2308.02490} {Mm-vet: Evaluating large multimodal models for integrated capabilities}.
\newblock \emph{arXiv preprint arXiv:2308.02490}.

\bibitem[{Yue et~al.(2024)Yue, Ni, Zhang, Zheng, Liu, Zhang, Stevens, Jiang, Ren, Sun, Wei, Yu, Yuan, Sun, Yin, Zheng, Yang, Liu, Huang, Sun, Su, and Chen}]{yue2024mmmu}
Xiang Yue, Yuansheng Ni, Kai Zhang, Tianyu Zheng, Ruoqi Liu, Ge~Zhang, Samuel Stevens, Dongfu Jiang, Weiming Ren, Yuxuan Sun, Cong Wei, Botao Yu, Ruibin Yuan, Renliang Sun, Ming Yin, Boyuan Zheng, Zhenzhu Yang, Yibo Liu, Wenhao Huang, and 3 others. 2024.
\newblock \href {https://arxiv.org/abs/2311.16502} {Mmmu: A massive multi-discipline multimodal understanding and reasoning benchmark for expert agi}.
\newblock \emph{Preprint}, arXiv:2311.16502.

\bibitem[{Zhang et~al.(2023)Zhang, Aljunied, Gao, Chia, and Bing}]{zhang2023m3exam}
Wenxuan Zhang, Mahani Aljunied, Chang Gao, Yew~Ken Chia, and Lidong Bing. 2023.
\newblock \href {https://proceedings.neurips.cc/paper_files/paper/2023/hash/2cc5aa6b3bb4c5ba2c4dcec95329ad3e-Abstract-Datasets_and_Benchmarks.html} {M3exam: A multilingual, multimodal, multilevel benchmark for examining large language models}.
\newblock \emph{Advances in Neural Information Processing Systems}, 36:5484--5505.

\end{thebibliography}

\appendices
\clearpage
\appendix
\appendixpage            
\addappheadtotoc         
\numberwithin{figure}{section}
\numberwithin{table}{section}

This appendix provides supplementary material to support the main findings of this work. It is organized as follows:

\begin{itemize}
    \item \textbf{\S\ref{app:related_work}: Literature Review} \\
    Reviews related work on cultural bias in LLMs and VLMs, multilingual and monolingual VQA datasets, and visual reasoning benchmarks.

    \item \textbf{\S\ref{app:guidelines}: Annotation Guidelines} \\
    Describes the iterative development of annotation guidelines, including domain definitions, question types, and quality control processes.

    \item \textbf{\S\ref{app:data_stats}: Data Statistics} \\
    Provides detailed statistics on annotated samples, cultural category distributions, and clean filtered samples per Arab country.

    \item \textbf{\S\ref{app:details}: Additional Technical Details} \\
    Includes prompt templates for generating augmented captions and Q\&A pairs, user interface screenshots, human evaluation examples, and evaluation score breakdowns.
   \item \textbf{\S\ref{app:fine_grained_analysis}: Fine-Grained Models Performance Analysis on Pearl-Lite by Country and Question Type} \\
Presents detailed heatmaps of model performance on the Pearl-Lite benchmark, with results broken down by country and question type.

\end{itemize}

\paragraph{Key Tables}
\begin{itemize}
    \item \textbf{Table~\ref{tab:question_types}:} Taxonomy of $11$ culturally-focused question types.
    \item \textbf{Table~\ref{tab:cultural_distribution}:} Image distribution across countries and cultural domains.
    \item \textbf{Table~\ref{appdx_table:examples}:} Evaluation scores from LVLM judges and human annotators.
    \item \textbf{Table~\ref{tab:shared_concepts_elemnts}:} A complete list of $61$ identified shared concepts is provided.
     \item \textbf{Table~\ref{appdx_tab:templete_examples}:} Examples of question templates for both multiple and single image prompts in Arabic, covering diverse reasoning types and cultural features.
    
\end{itemize}

\paragraph{Key Figures}

\begin{figure*}
    \centering
    \includegraphics[width=0.99\linewidth, trim={0 0 0 9}, clip]{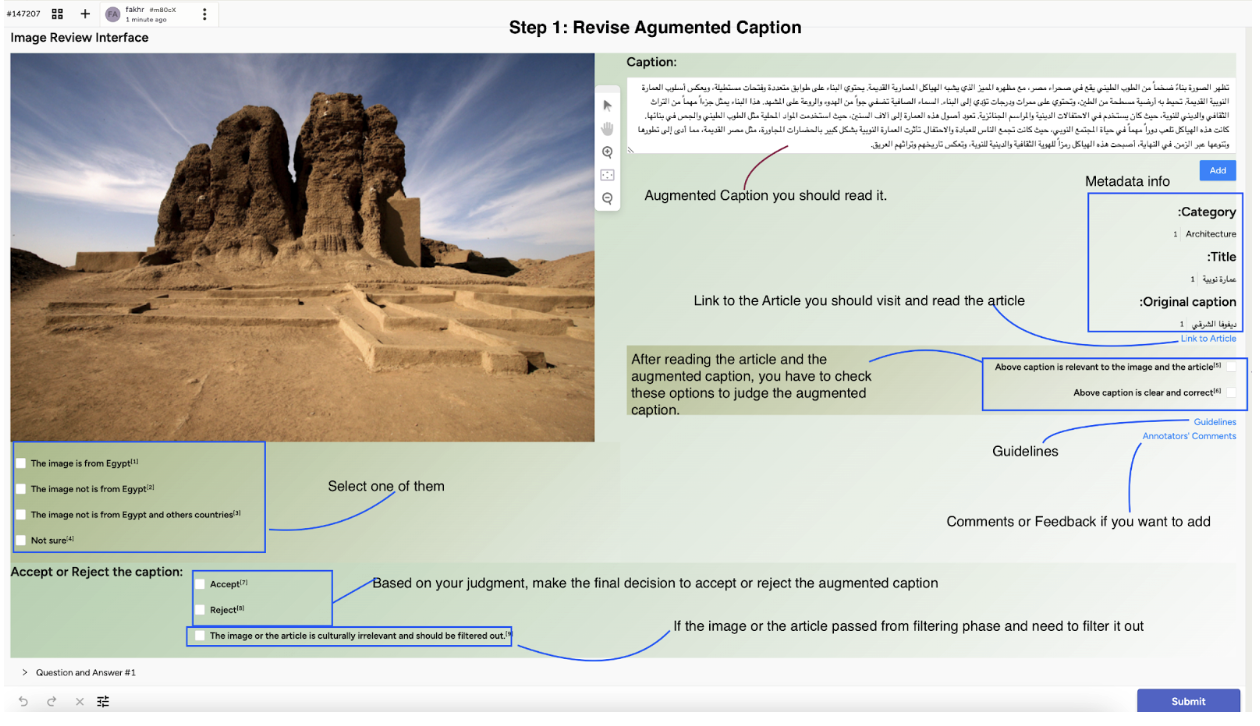}
    \caption{Label Studio platform interface used by annotators in step one of the human revision phase to revise the augmented caption}
    \label{fig:platform_example}
\end{figure*}

\begin{figure*}
    \centering
    \includegraphics[width=0.99\linewidth]{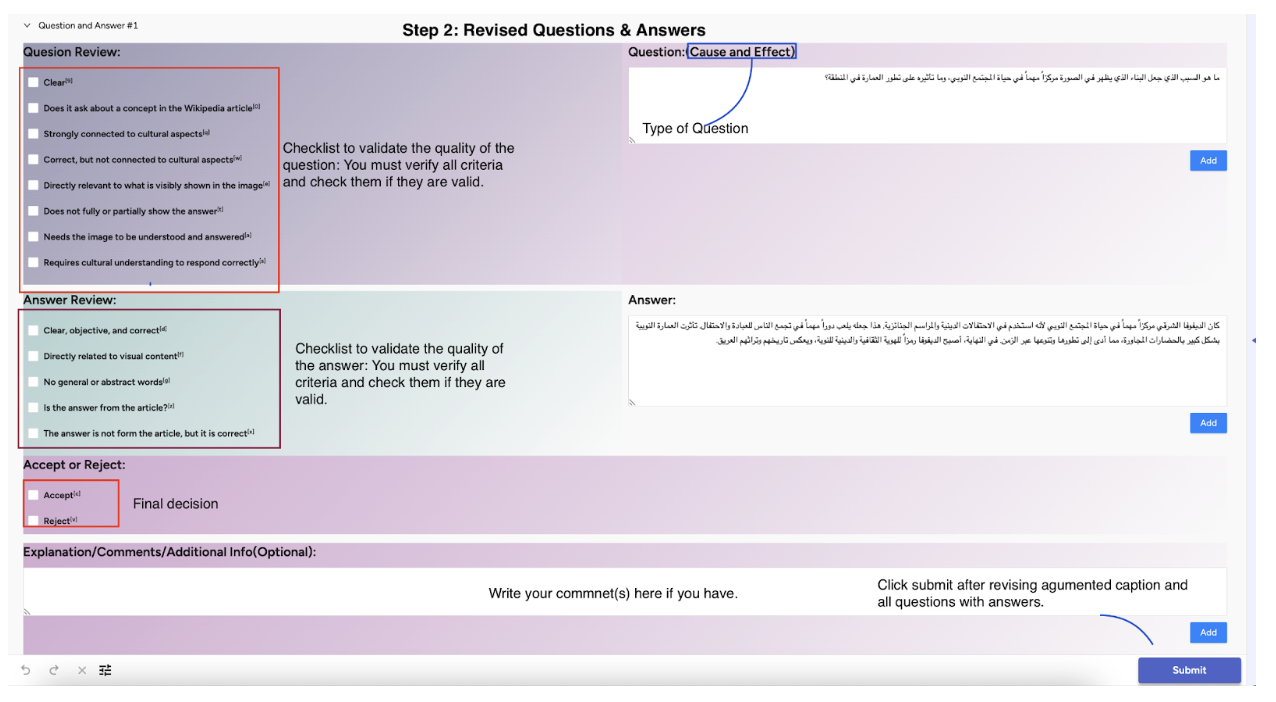}
    \caption{Step two of the revision phase, showing the process of reviewing and editing questions and answers on the platform}
    \label{fig:platform_step2}
\end{figure*}

\begin{figure*}
    \centering
    \includegraphics[width=0.99\linewidth]{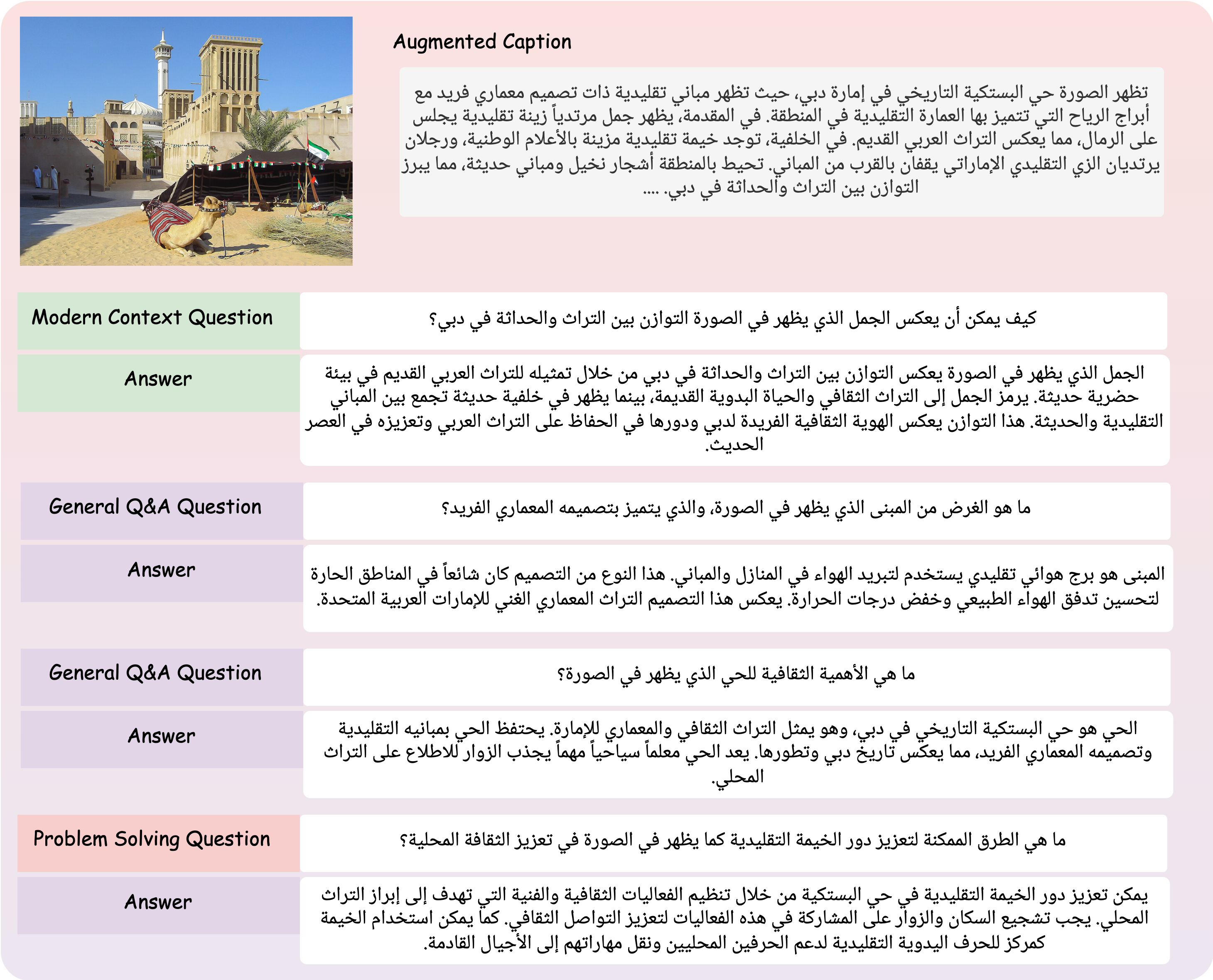}
    \caption{Shows an example illustrates the format of the augmented caption followed by multiple Q\&A pairs. Each question is linked to a specific reasoning type and is designed to reflect cultural understanding grounded in the image and caption.}
    \label{fig:dubai_example}
\end{figure*}

\begin{figure*}
    \centering
    \includegraphics[width=0.99\linewidth]{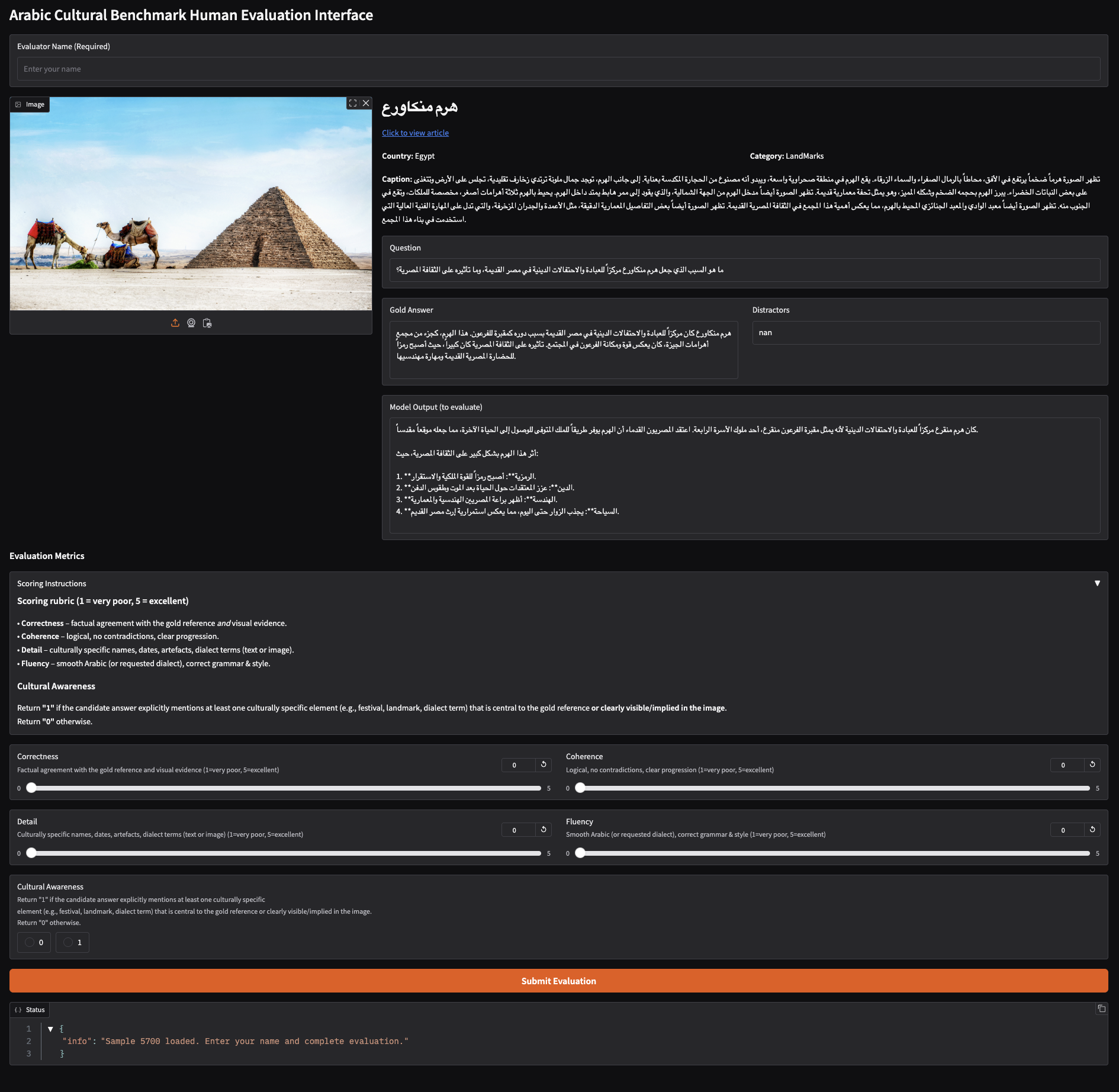}
    \caption{Displays User interface employed by annotators during the human evaluation phase, where they rate the models’ output based on Correctness, Coherence, Detail, and Fluency}
    \label{fig:eval_screen}
\end{figure*}

 \begin{itemize}
    \item \textbf{Figure~\ref{tab:cultural_distribution}:} Distribution of the number of images by cultural category for each Arab country in our dataset.

      \item \textbf{Figure~\ref{appdx_fig:prompt_step1}:} Shows prompt template step 1 used to generate augmented captions.
      \item \textbf{Figure~\ref{appdx_fig:prompt_step2/part1}:} Illustrates template step 2 (Part 1) to generate Q\&A pairs.
      \item \textbf{Figure~\ref{appdx_fig:prompt_step2/part2}:} Displays prompt template step 2 (Part 2) used to generate answers.
      \item \textbf{Figure~\ref{appdx_fig:prompt_combined}:} Combined prompt template including the two steps for generating Q\&A pairs.
      \item \textbf{Figure~\ref{fig:platform_example}:} Label Studio platform interface used by annotators in step one of the human revision phase to revise the augmented caption.
      \item \textbf{Figure~\ref{fig:platform_step2}:} Step two of the revision phase, showing the process of reviewing and editing questions and answers on the platform.
      \item \textbf{Figure~\ref{fig:dubai_example}:} Shows an example illustrates the format of the augmented caption followed by multiple Q\&A pairs.
      \item \textbf{Figure~\ref{fig:eval_screen}:} Human Eval UI
      \item \textbf{Figure~\ref{appdx_fig:prompt_eval}:} Evaluation prompts examples
      \item \textbf{Figure~\ref{appdx_fig:prompt_new}:} Illustrates the prompt used to generate Q\&A pairs for \textit{Pearl X} using \texttt{OpenAI o3}.

      \item \textbf{Figure~\ref{fig:accuracy_by_model_country}:} Heatmap of accuracy scores on closed-form questions by model and country.
      \item \textbf{Figure~\ref{fig:overall_by_model_country}:} Heatmap of the Overall Score for open-ended questions by model and country.
      \item \textbf{Figure~\ref{fig:cas_by_model_country}:} Heatmap of the CAS for open-ended questions by model and country.
     \item \textbf{Figure~\ref{fig:overall_by_type_country}:} Heatmap of the Overall Score for open-ended questions by model and question type.

\end{itemize}

\section{Literature Review}
\label{app:related_work}
\vspace{1mm}\noindent\textbf{Culture Bias in LLMs and LVMs. }
Beyond language understanding capabilities, cultural awareness and sensibility are critical to ensure practical effectiveness while mitigating the potential stereotypes and biases of text-based and multimodal LLMs. Thus, designing culturally aware LLMs requires understanding the various perspectives of a given culture. This involves recognizing diverse cultural dimensions such as traditions, beliefs, and social practices as one dimension, social interaction as the second dimension, and materialized objects as the third dimension \cite{pawar2024survey}. Hence, different studies have been conducted to evaluate and identify cultural gaps and ensure diversity and inclusion in current SOTA LLMs.  

\cite{pawar2024survey} provided a comprehensive survey on recent works on cultural awareness, exploring dataset creation methodologies, benchmarking techniques, and the ethical implications. For example, one of the main methodologies for dataset creation is relying on automatic pipelines that leverage public corpora to generate large datasets \cite{sengupta2023jais,huang2024acegpt,aloui2024101}. However, incorporating information from a specific culture into a general-purpose LLM can lead to misinformation, stereotyping, biases, and misrepresentation when used to represent other cultures \cite{pawar2024survey}. Generally, this stems from the use of machine-translated data to create multilingual LLMs, a widely adopted practice in the field \cite{hendrycks2020measuring,singh2024global}. In particular, \cite{singh2024global} highlighted the propagated emphasis of Western perspectives on the topics covered in the translated English dataset, such as the MMLU benchmarks \cite{hendrycks2020measuring}. 

To create a culturally aligned dataset, other works employ humans either as part of semi-automatic methods or completely to manually create cultural datasets from scratch \cite{pawar2024survey,ma2023food,alyafeai2024cidar,huang2023acegpt,baek2024evaluating}. For example, to ensure a multilingual culturally diverse dataset, \cite{singh2024global} introduced Global-MMLU, an improved set validated by humans and covering culturally sensitive and agnostic sets of $42$ languages. 

Similarly, relying on English pre-trained LLMs to build VLMs is the core cause of inherently encoding Western cultural knowledge. Thus, most vision-language models exhibit cultural misrepresentation \cite{burda2024culturally,ananthram2024see}. Images often convey rich cultural stories and heritages; however, English-based VLM-generated captions tend to fail to accurately narrate cultural stories \cite{burda2024culturally}. To study this phoneme, several works have been conducted to assess the performance of VLMs in culturally specific information \cite{burda2024culturally,ananthram2024see,bhatia2024local}.

For example, \cite{burda2024culturally} assessed the performance of four VLMs' capabilities to identify cultural information in images to generate aligned culturally sensitive captions. Most of the models struggle to perform the task, as the highest cultural awareness score of 35\% was achieved by Gemini Pro Vision. Furthermore, \cite{bhatia2024local} introduced GLOBALRG benchmark to evaluate multicultural understanding and inclusivity in VLMs across universal and local concepts. Although the results of evaluating several SOTA VLMs on the image retrieval task varied across cultures, most of the retrieved images contain Western-specific elements. To mitigate the interpretation of images from the Western perspective only, \cite{ananthram2024see} utilized a mix of diverse language-based VLMs to improve the model's understanding ability of Chinese cultural images.

\vspace{1mm}\noindent\textbf{Monolingual VQA Datasets. }
Several studies have addressed the VQA problem by developing language-specific datasets. Datasets like MMT-Bench \cite{ying2024mmtbench}, CAMEL-Bench \cite{ghaboura2024camel}, and JMMMU \cite{onohara2024jmmmu} considered only the MCQ type for English, Arabic, and Japanese languages, respectively. Similarly, 
the HaVQA \cite{parida-etal-2023-havqa} and ViTextVQA \cite{vannguyen2024ViTextVQA} datasets use open-ended questions in Hausa and Vietnamese languages, respectively. VLBiasBench \cite{wang2024vlbiasbench} offers a large English dataset with both open and closed questions. A semi-automated method was used to create K-Viscuit \cite{baek2024evaluating}, a Korean cultural dataset generated with GPT-4 and human verification. 
In Arabic VQA, CAMEL-Bench \cite{ghaboura2024camel} includes $29,036$ curated MCQs across various domains. This dataset integrates Arabic and translated content from existing English LMM benchmarks. It mainly focuses on Modern Standard Arabic (MSA) with limited attention to the Arabic dialects.

\vspace{1mm}\noindent\textbf{Visual Reasoning. }
Reasoning represents one of the fundamental human cognitive abilities derived from commonsense understanding and world knowledge. The cognitive process involves interpreting information, analyzing given conditions, and subsequently drawing inferences, solving problems, or making predictions. Although Multimodal LLMs (MLLMs) have achieved remarkable advancements across various domains, their capacity for robust reasoning, particularly in processing and integrating information from diverse modalities, remains a prominent and challenging research frontier. Consequently, a growing body of research has focused on evaluating and improving the reasoning capabilities of MLLMs, resulting in the development of specialized datasets and benchmarks aimed at rigorously assessing models' reasoning capabilities. 

Several benchmarks have been proposed to evaluate different facets of MLLMs reasoning. InfiMM-Eval benchmark \cite{han2023infimm} was designed to evaluate the reasoning capabilities of MLLMs through open-ended and multi-step reasoning across three categories: deductive, abductive, and analogical. This benchmark consists of $279$ manually curated, high-quality questions paired with $342$ images, placing a strong emphasis on the multi-step reasoning process. 
Further contributing to the evaluation and development of reasoning in MLLMs, the Large-scale Visual Chain-of-Thought (Visual CoT) dataset \cite{NEURIPS2024_0ff38d72} aims to improve the inheritability and reasoning abilities of MLLMs. It contains $438$K visual question-answer pairs annotated with intermediate bounding boxes that highlight critical image regions. Around $98$K of these pairs are further enriched with explicit CoT annotations that provide detailed reasoning steps. 

Focusing specifically on spatial understanding, the SpatialEval dataset \cite{wang2024picture} covers four spatial reasoning tasks, such as spatial relationships, navigation, positional understanding, and counting, to evaluate the spatial reasoning capabilities of LLMs and VLMs.
Focusing on the same facet of reasoning, the SpatialVLM framework \cite{Chen_2024_CVPR}  introduces an approach to enhance the spatial reasoning abilities of VLMs by generating large-scale 3D spatial reasoning data. This framework transforms 2D images into detailed metric-scale 3D point cloud, enabling the synthesis of approximately two billion spatial reasoning QA pairs. These pairs are designed to cover both qualitative and quantitative spatial reasoning tasks.

CogBench \cite{song2025} is a cognitive evaluation benchmark designed to assess the reasoning abilities of LVLMs using a unique set of images inspired by the "Cookie Theft" cognitive assessment task. The dataset consists of manually collected images, mostly painting-style from Pinterest, which then underwent detailed human annotation. Annotators identified key entities, constructed explicit Chain-of-Reasoning (CoR) annotations, and provided thorough descriptions of the image content. 
This results in $251$ annotated images, structured across eight reasoning dimensions and includes two primary evaluation tasks: an image description task and a multiple-choice VQA task.
Similarly, the MM-Vet benchmark \cite{yu2023mm} has been designed to evaluate LMMs on complex multimodal tasks requiring integrated vision language capabilities. The benchmark covers six core VL abilities, including recognition, OCR, knowledge, spatial awareness, language generation, and arithmetic capability. 

Despite these advancements, the development of comprehensive evaluation datasets remains crucial. A systematic review by \cite {wang2024exploring} highlighted the current State-of-the-art reasoning capabilities within MLLMs and noted that while various datasets have been proposed as multimodal reasoning benchmarks, many of which lack comprehensive reasoning steps. This underscores the importance of developing robust benchmarks that can accurately measure and drive progress in the multimodal reasoning capabilities of current models.

\section{Annotation Guidelines}
We iteratively developed the annotation guidelines for \ourdataset over a period of six months, structured across two distinct phases. In \textit{Phase I}, annotators focused on filtering \textit{Wikipedia} articles based on two main criteria: cultural uniqueness and the relevance of images to their corresponding articles. Throughout this initial phase, annotators provided continuous feedback during weekly meetings, allowing us to promptly address and resolve any encountered issues. The annotation guidelines begin by outlining the primary goals of the project, followed by detailed descriptions of the ten domains, each supported by illustrative images and captions to ensure clarity. The second part of the guidelines explains each question type used in the project, providing clear definitions along with practical examples directly from the annotation platform to support annotators throughout the process.\\
\textit{Phase III} involved reviewing automatically generated questions derived from the previously filtered data. Annotators were responsible for thoroughly validating these questions and ensuring the accuracy and appropriateness of question types and answers. Throughout this second phase, we further refined and enhanced the annotation guidelines based on annotators' experiences and feedback, ensuring consistency and effectiveness across all phases of the annotation process.
\label{app:guidelines}
\begin{table*}[h]
\centering
\renewcommand{\arraystretch}{1.3}  
\scalebox{0.68}{%
\begin{tabular}{p{4cm} p{2cm} p{13cm}}
\toprule
\rowcolor{gray!30}
\textbf{Question Type} & \textbf{Count} & \textbf{Purpose} \\
\midrule
\textit{Cause and Effect}       & $23,400$ & Ask about the reasons behind a cultural element and its consequences or impacts. \\
\textit{Chronological Sequence}  & $22,614$ & Examine the historical development or timeline of a cultural element or practice. \\
\textit{Comparative Analysis}    & $23,264$ & Compare or contrast cultural elements within or across contexts or regions. \\
\textit{General Q\&A}           & $35,034$ & Query a straightforward factual detail about the image or topic as stated in the article. \\
\textit{Hypothesis Formation}    & $23,552$ & Pose a speculative why/how question about a cultural element, prompting an explanatory theory. \\
\textit{Modern Context}          & $22,972$ & Connect a traditional cultural element to its relevance or adaptation in today's world. \\
\textit{Origin Identification}   & $23,776$ & Inquire about the historical or geographical origin of a cultural element or practice. \\
\textit{Perspective Shifting}    & $23,498$ & Explore different viewpoints or interpretations regarding the cultural element. \\
\textit{Problem Solving}        & $23,764$ & Present a cultural challenge or issue and ask for a solution or mitigation approach. \\
\textit{Role Playing}           & $23,184$ & Provide an answer from a specific role or persona related to the cultural context. \\
\textit{Scenario Completion}     & $23,368$ & Present an incomplete scenario or sequence and ask to predict or complete the outcome. \\
\textit{Multiple Choice}        & $30,654$ & Present several options for a question, requiring selection of the correct answer. \\
\textit{True/False}             & $10,218$ & Present a statement and ask whether it is true or false, testing factual accuracy of cultural information. \\
\bottomrule
\end{tabular}
}
\caption{Lists the 13 question types used in \ourdataset, with their counts in Pearl and intended purposes, each designed to elicit different forms of cultural reasoning.}
\label{tab:question_types}
\end{table*}

\begin{figure*}
    \centering
    \includegraphics[width=0.90\linewidth]{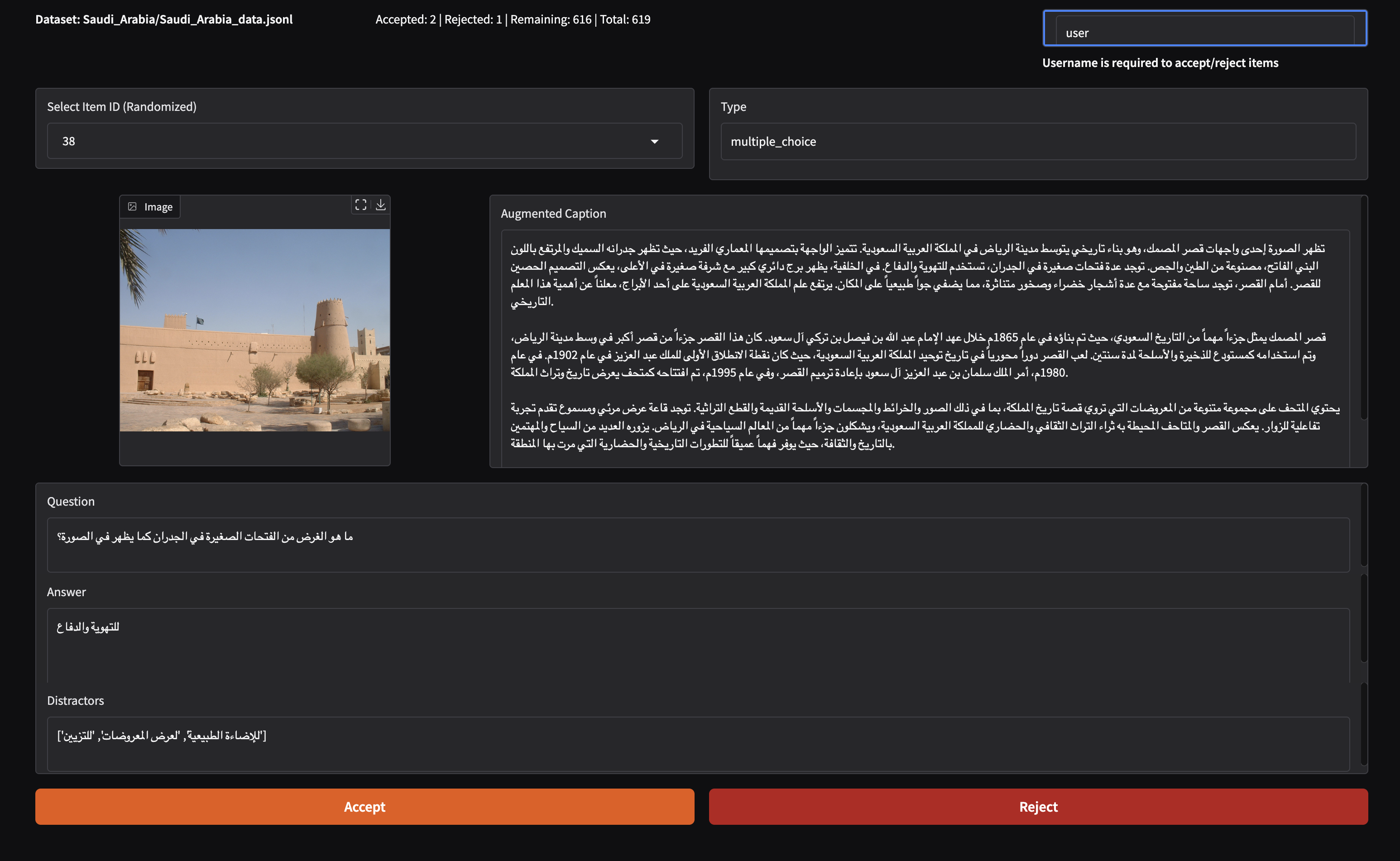}
    \caption{User Interface used by annotators to revise $11$K Q\&A pairs for benchmarking. The revision process involved filtering samples based on cultural relevance and quality, enabling annotators to accept or reject entries accordingly.}
    \label{fig:benchmark_revision_ui}
\end{figure*}

\section{Data Statistics}
\label{app:data_stats}
- We provide statistics on the number of images by cultural category for the Arab countries in \ourdataset in ~\ref{tab:cultural_distribution}.\\

\begin{table*}[htbp]
  \centering
  \rowcolors{2}{gray!15}{white}
\scalebox{0.6}{%

  \begin{tabular}{@{}lrrrrrrrrrrr@{}}
    \toprule
    Country        & Architecture & Clothes & Fauna & \makecell{Festivals \&\\Celebrations} & Flora & Food & Geography & Handicrafts & Landmarks & Music & Total \\
    \midrule
    Algeria        &     $0$ &  $76$  &   $0$ &   $0$ &   $0$ &   $0$ &    $0$ &    $1$ &   $32$ & $143$ &  $252$  \\
    Bahrain        &     $0$ &   $0$  &   $1$ &   $0$ &   $0$ &  $16$ &  $143$ &    $1$ &    $0$ &  $17$ &  $178$  \\
    Egypt          &   $4,63$ &   $1$  &   $6$ &   $1$ &  $12$ & $162$ &  $211$ &    $0$ &  $264$ & $252$ & $1,372$  \\
    Iraq           &   $170$ &   $3$  &  $14$ &   $0$ &   $2$ &  $62$ &   $74$ &    $0$ &    $0$ &  $85$ &  $410$  \\
    Jordan         &   $400$ &   $0$  & $145$ &   $0$ &   $7$ &  $60$ & $1,546$ &    $0$ &  $982$ &  $25$ & $3,165$  \\
    Kuwait         &   $111$ &   $0$  &   $7$ &   $0$ &  $10$ &   $0$ &  $212$ &    $0$ &    $0$ &  $65$ &  $405$  \\
    Lebanon        &   $117$ &   $0$  &  $41$ &   $0$ & $101$ &  $66$ &  $280$ &    $0$ &    $0$ & $191$ &  $796$  \\
    Libya          &    $10$ &   $0$  &  $20$ &   $0$ &  $53$ &  $22$ &  $296$ &    $0$ &  $155$ &  $10$ &  $566$  \\
    Mauritania     &     $0$ &   $0$  &   $2$ &   $0$ &  $18$ &   $7$ &  $198$ &    $0$ &    $0$ &   $0$ &  $225$  \\
    Morocco        &   $223$ & $105$  &  $52$ &   $1$ & $203$ & $235$ &  $310$ &  $133$ &  $123$ & $256$ & $1,641$  \\
    Oman           &   $141$ &  $12$  &  $55$ &   $0$ &  $36$ &   $4$ &  $331$ &    $0$ &    $0$ &   $5$ &  $584$  \\
    Palestine      &  $1,091$ &  $93$  &  $73$ &   $0$ & $185$ & $393$ &   $76$ &   $34$ &  $294$ &  $48$ & $2,287$  \\
    Qatar          &     $0$ &   $0$  &  $13$ &   $0$ &   $0$ &   $1$ &  $144$ &    $0$ &    $0$ &   $4$ &  $162$  \\
    Saudi Arabia   &   $220$ &  $12$  & $121$ &  $14$ & $111$ &  $33$ &  $356$ &    $0$ &    $0$ &  $22$ &  $889$  \\
    Sudan          &     $0$ &   $0$  &   $1$ &   $0$ &   $0$ &  $16$ &  $339$ &    $0$ &  $113$ &   $0$ &  $469$  \\
    Syria          &   $877$ &  $11$  &   $8$ &   $0$ & $134$ &  $61$ &  $253$ &    $2$ &    $0$ &   $7$ & $1353$  \\
    Tunisia        &    $92$ &  $11$  &   $5$ &   $0$ &   $0$ &   $4$ &    $0$ &    $4$ &    $0$ &   $0$ &  $116$  \\
    UAE            &     $4$ &   $0$  &   $7$ &   $0$ &   $0$ &   $0$ &  $135$ &    $0$ &    $0$ &   $3$ &  $149$  \\
    Yemen          &   $385$ &   $2$  &   $0$ &   $0$ &  $42$ &  $63$ &  $551$ &    $0$ &    $0$ &  $23$ & $1,066$  \\
    \midrule
    \textbf{Total} &  $4,304$ & $326$  & $571$ &  $16$ & $914$ &$1,205$ & $5,455$ &  $175$ & $1,963$ &$1,156$ &$16,085$  \\
    \bottomrule
  \end{tabular}
  }
  \caption{Distribution of the number of images by cultural category for each Arab country in \ourdataset (after revision).}
  \label{tab:cultural_distribution}
\end{table*}

\section{Additional Technical Details}
\label{app:details}

\begin{figure*}[ht]  
\centering
\begin{tcolorbox}[
    title={\bfseries Prompt},
    width=\textwidth,
    colframe=gray!75!black,
    colback=blue!5,
    colbacktitle=gray!75!black,
    coltitle=white,
    boxsep=3pt,
    arc=1mm,
    fontupper=\small\ttfamily,
    before upper={\setlength{\parskip}{0.5em}\setlength{\baselineskip}{0.9\baselineskip}},
]
{\color{blue!80!black}\bfseries Step 1: Generate an Augmented Caption}

Your first task is to create a detailed, extended description of the image based on the provided image caption and Wikipedia article. This augmented caption should:

{\setlength{\parskip}{0.2em}
\begin{enumerate}
\setlength{\itemsep}{0pt}
\setlength{\parsep}{0pt}
\item START with $\RL{تظهر الصورة}$ (The image shows) followed by a comprehensive description of what is VISUALLY present in the image, using the provided image caption as a foundation.
\item Expand on the visual elements by adding relevant contextual information from the provided sources.
\item Focus particularly on details that would support \{focus\_area\} question-answer pairs.
\item Describe the image's visual elements in detail including:
   \begin{itemize}
   \setlength{\itemsep}{0pt}
   \setlength{\parsep}{0pt}
   \item Specific objects, people, settings, activities, and artifacts visible
   \item Spatial relationships between elements
   \item Notable colors, textures, and visual features
   \item Any text or inscriptions visible
   \item Any actions or events being depicted
   \end{itemize}
\end{enumerate}}

{\color{blue!80!black}\bfseries INFORMATION GUIDELINES:}

{\setlength{\parskip}{0.2em}
\begin{enumerate}
\setlength{\itemsep}{0pt}
\setlength{\parsep}{0pt}
\item Include information from the provided sources (category, title, country, image caption, Wikipedia article)
\item Present information as factual statements without attributing to the source
\item DO NOT use phrases like ``according to the article,'' ``as mentioned in the caption,'' or any reference to the sources
\item DO NOT mention Wikipedia, articles, captions, or sources in any way
\item Simply state the facts and information directly as established knowledge
\end{enumerate}}

{\color{blue!80!black}\bfseries Question Type-Specific Requirements for \{question\_type\}:}

\{specific\_requirements\}

{\color{blue!80!black}\bfseries Input Information}

{\setlength{\parskip}{0.2em}
\begin{itemize}
\setlength{\itemsep}{0pt}
\setlength{\parsep}{0pt}
\item {\bfseries Cultural Category:} \{category\}
\item {\bfseries Article Title:} \{title\}  
\item {\bfseries Country:} \{country\}
\item {\bfseries Image Caption:} \{image\_caption\}
\item {\bfseries Wikipedia Article:} \{wikipedia\_article\}
\end{itemize}}

{\color{blue!80!black}\bfseries REQUIRED STRUCTURE:}

{\setlength{\parskip}{0.2em}
\begin{enumerate}
\setlength{\itemsep}{0pt}
\setlength{\parsep}{0pt}
\item First paragraph: Begin with ``\RL{تظهر الصورة}'' and describe what is VISUALLY present based on the image caption
\item Following paragraphs: Add relevant cultural context from the sources that directly relates to the visual elements
\item Final paragraph: Summarize elements specifically relevant to \{question\_type\} questions
\end{enumerate}}

{\color{blue!80!black}\bfseries CRITICAL VERIFICATION STEP:}

Before finalizing your augmented caption, verify that:
{\setlength{\parskip}{0.2em}
\begin{itemize}
\setlength{\itemsep}{0pt}
\setlength{\parsep}{0pt}
\item The caption has sufficient detail to support \{question\_type\} question-answer pairs
\item You have NOT included any reference to articles, captions, or sources
\item You have presented all information directly as factual statements
\end{itemize}}

If the provided sources do not provide SUFFICIENT information to create a meaningful augmented caption for \{question\_type\} question-answer pairs, return a JSON error message:

\begin{tcolorbox}[colback=gray!10, colframe=gray!40, size=fbox, boxrule=0.5pt, left=1pt, right=1pt, top=1pt, bottom=1pt]
\footnotesize
\ttfamily
\{"error": "Insufficient context in the provided sources to generate an augmented caption for \{question\_type\} question-answer pairs."\}
\end{tcolorbox}
\end{tcolorbox}
\caption{Shows \textit{prompt template step 1} used to generate augmented captions}
\label{appdx_fig:prompt_step1}
\end{figure*}

\begin{figure*}[ht]  
\centering
\begin{tcolorbox}[
    title={\bfseries Prompt (Step 2: Part 1)},
    width=\textwidth,
    colframe=gray!75!black,
    colback=black!5,
    colbacktitle=gray!75!black,
    coltitle=white,
    boxsep=3pt,
    arc=1mm,
    fontupper=\small\ttfamily,
    before upper={\setlength{\parskip}{0.3em}\setlength{\baselineskip}{0.85\baselineskip}},
    breakable=false
]
{\color{blue!80!black}\bfseries Step 2: Generate Question-Answer Pairs}

Using ONLY the augmented caption you created in Step 1, now generate \{num\_pairs\} different question-answer pairs that \{task\_description\}, without explicitly naming it.

All questions and answers MUST be written in Modern Standard Arabic only.

{\color{blue!80!black}\bfseries INFORMATION GUIDELINES:}

{\setlength{\parskip}{0.2em}
\begin{enumerate}
\setlength{\itemsep}{0pt}
\setlength{\parsep}{0pt}
\item Questions and answers MUST be based EXCLUSIVELY on information in the augmented caption
\item DO NOT introduce any new information not present in the augmented caption
\item DO NOT mention or reference any articles, captions, or sources
\item Present all information as direct factual statements without attribution
\item If the augmented caption lacks sufficient information for meaningful Q\&A pairs, return an error
\end{enumerate}}

{\color{blue!80!black}\bfseries Question Requirements:}

{\setlength{\parskip}{0.2em}
\begin{enumerate}
\setlength{\itemsep}{0pt}
\setlength{\parsep}{0pt}
\item Your question MUST reference something clearly described in the augmented caption
\item Your question MUST use one of these exact phrases to refer to the element:
   \begin{itemize}
   \setlength{\itemsep}{0pt}
   \setlength{\parsep}{0pt}
   \item ``\RL{الذي يظهر في الصورة}'' (that appears in the image)
   \item ``\RL{كما يظهر في الصورة}'' (as shown in the image)
   \item ``\RL{الظاهر في الصورة}'' (the visible element in the image)
   \end{itemize}
\item NEVER mention the specific name of the element in the question
\item Each question should require both visual identification AND cultural knowledge
\item NEVER include any terms that could hint at the exact name of the object, tradition, landmark, or feature
\end{enumerate}}
\end{tcolorbox}
\caption{Illustrates template step 2 (Part 1) to generate Q\&A pairs}
\label{appdx_fig:prompt_step2/part1}
\end{figure*}

\begin{figure*}[ht]  
\centering
\begin{tcolorbox}[
    title={\bfseries Prompt (Step 2: Part 2)},
    width=\textwidth,
    colframe=gray!75!black,
    colback=blue!5,
    colbacktitle=gray!75!black,
    coltitle=white,
    boxsep=3pt,
    arc=1mm,
    fontupper=\small\ttfamily,
    before upper={\setlength{\parskip}{0.3em}\setlength{\baselineskip}{0.85\baselineskip}},
    breakable=false
]
{\color{blue!80!black}\bfseries Answer Requirements:}

{\setlength{\parskip}{0.2em}
\begin{enumerate}
\setlength{\itemsep}{0pt}
\setlength{\parsep}{0pt}
\item Base all answers EXCLUSIVELY on information in the augmented caption
\item NEVER add any details or context not in the augmented caption
\item Keep answers between 2-5 sentences in length
\item The answer MUST directly address and resolve the specific question being asked
\item Start your answer with a direct response to the question, then provide supporting details
\item DO explicitly name the object, tradition, or element in your answer
\item You SHOULD include the specific name of elements in the answer - unlike in the question
\item Use clear language and structured sentences that directly connect to the question
\item AVOID repeating the same information across multiple answers
\item Include the country/region name ONLY when it is relevant to the answer
\item NEVER mention or reference articles, captions, sources, Wikipedia, or any attribution phrases
\end{enumerate}}

{\color{blue!80!black}\bfseries Question Type-Specific Requirements for \{question\_type\}:}

\{specific\_requirements\}

{\color{blue!80!black}\bfseries CRITICAL VERIFICATION STEP:}

Before finalizing each Q\&A pair, verify that:
{\setlength{\parskip}{0.2em}
\begin{itemize}
\setlength{\itemsep}{0pt}
\setlength{\parsep}{0pt}
\item The answer contains information found ONLY in the augmented caption
\item No new information has been introduced
\item The question uses one of the required phrases
\item The question doesn't name the specific element
\item The answer DOES directly name the specific element
\item The answer directly and clearly addresses the specific question being asked
\item The answer follows the required length guidelines (2-5 sentences)
\item NO reference is made to any sources such as articles or captions
\end{itemize}}

If the augmented caption does not provide SUFFICIENT information to create meaningful question-answer pairs, return a JSON error message:

\begin{tcolorbox}[colback=gray!10, colframe=gray!40, size=fbox, boxrule=0.5pt, left=1pt, right=1pt, top=1pt, bottom=1pt]
\footnotesize
\ttfamily
\{"error": "Insufficient context in the augmented caption to generate meaningful question-answer pairs for this question type."\}
\end{tcolorbox}
\end{tcolorbox}
\caption{Displays prompt template step 2 (Part 2) used to generate answers}
\label{appdx_fig:prompt_step2/part2}
\end{figure*}

\begin{figure*}[ht]  
\centering
\begin{tcolorbox}[
    title={\bfseries Prompt},
    width=\textwidth,
    colframe=gray!75!black,
    colback=blue!5,
    colbacktitle=gray!75!black,
    coltitle=white,
    boxsep=3pt,
    arc=1mm,
    fontupper=\small\ttfamily,
    before upper={\setlength{\parskip}{0.5em}\setlength{\baselineskip}{0.9\baselineskip}},
]
{\color{blue!80!black}\bfseries Two-Step Process for Generating Question-Answer Pairs}

\{step1\_prompt\}

\{step2\_prompt\}

\{example\}

{\color{blue!80!black}\bfseries Final Output Format}
Your final output must be a valid JSON object with the following structure:

\begin{tcolorbox}[colback=gray!10, colframe=gray!40, size=fbox, boxrule=0.5pt, left=1pt, right=1pt, top=1pt, bottom=1pt]
\footnotesize
\ttfamily
\{\\
  "augmented\_caption": "The generated augmented caption based on the question-answer type.",\\
  "generated\_QAs": [\\
    \{\\
      "question": "Your first question text in Modern Standard Arabic here",\\
      "answer": "Your first answer text in Modern Standard Arabic here"\\
    \},\\
    \{\\
      "question": "Your second question text in Modern Standard Arabic here",\\
      "answer": "Your second answer text in Modern Standard Arabic here"\\
    \}\\
    // Additional pairs as needed up to \{num\_pairs\}...\\
  ]\\
\}
\end{tcolorbox}

If there is insufficient context in the provided sources for either step, your output should be a JSON object with an error message:

\begin{tcolorbox}[colback=gray!10, colframe=gray!40, size=fbox, boxrule=0.5pt, left=1pt, right=1pt, top=1pt, bottom=1pt]
\footnotesize
\ttfamily
\{"error": "Specific error message explaining the lack of sufficient context."\}
\end{tcolorbox}

{\color{blue!80!black}\bfseries FINAL REMINDER - CRITICAL INSTRUCTIONS:}

{\setlength{\parskip}{0.2em}
\begin{itemize}
\setlength{\itemsep}{0pt}
\setlength{\parsep}{0pt}
\item The augmented caption MUST begin with ``\RL{تظهر الصورة}'' followed by a detailed description of what is VISUALLY present
\item ALL information must be based on the provided sources
\item Present ALL information as direct factual statements
\item DO NOT use phrases like ``according to the article,'' ``as mentioned,'' or similar attributions
\item DO NOT reference Wikipedia, articles, captions, or any sources in the augmented caption, questions, or answers
\item Questions should NOT mention the specific name of elements, but answers SHOULD explicitly name them
\item Answers MUST directly respond to their corresponding questions and be between 2-5 sentences in length
\item Each answer should begin with a sentence that directly addresses the question asked
\item If you cannot generate meaningful content without inventing information, return an error message
\end{itemize}}
\end{tcolorbox}
\caption{Combined prompt template including the two steps for generating Q\&A pairs}
\label{appdx_fig:prompt_combined}
\end{figure*}

\begin{table*}[]
\resizebox{\textwidth}{!}{%
\begin{tabular}{rllrl}
\toprule
\multicolumn{1}{c}{\textbf{Image}} & \textbf{Question Type}                   & \multicolumn{2}{c}{\textbf{Text}}                                                                                                                                                                & \textbf{Translation}                                                                                                                                                                                                          \\ \midrule
\multirow{2}{*}{\adjustbox{valign=m}{\includegraphics[width=1.9cm]{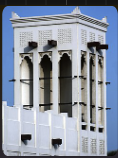}}}                        & \multirow{2}{*}{Hypothesis formation}    & Caption  & \begin{tabular}[c]{@{}r@{}}\RL{ تظهر الصورة برج الرياح التقليدي في البحرين، وهو عنصر معماري }\\ \RL{ تراثي يستخدم لتبريد المنازل قديما قبل ظهور التكييف الكهربائي }\end{tabular}   & \begin{tabular}[c]{@{}l@{}}The picture shows a traditional wind tower in Bahrain—an architectural\\ heritage feature that was once used to cool homes before\\ the advent of electric air-conditioning.\end{tabular}          \\ \cmidrule{3-5}
                                          &                                          & Question & \begin{tabular}[c]{@{}r@{}}\RL{ لماذا تعتقد أن المعماريين في البحرين والخليج العربي اخترعوا البرج }\\ \RL{ الذي يظهر في الصورة؟ }\end{tabular}                                        & \begin{tabular}[c]{@{}l@{}}Why do you think architects in Bahrain and the Gulf\\ region invented the tower shown in the picture?\end{tabular}                                                                                 \\ \midrule
\multirow{2}{*}{\adjustbox{valign=m}{\includegraphics[width=2.5cm]{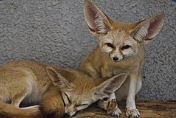}}}                        & \multirow{2}{*}{Hypothesis formation}    & Caption  & \begin{tabular}[c]{@{}r@{}}\RL{ تظهر الصورة زوجا من حيوانات الفنك، يتميزان بفرائهما الرملي الكثيف }\\ \RL{ وآذانهما الكبيرة البارزة، وهما مستلقيان }\end{tabular}                   & \begin{tabular}[c]{@{}l@{}}The image depicts a pair of fennec foxes, distinguished by\\ their thick sandy fur and large, prominent ears, reclining side\\ by side.\end{tabular}                                               \\ \cmidrule{3-5}
                                          &                                          & Question & \begin{tabular}[c]{@{}r@{}}\RL{ لماذا تعتقد أن الفنك أصبح رمزا ثقافيا مهما في الجزائر، }\\ \RL{ ويستخدم في الرياضة؟ }\end{tabular}                                                & \begin{tabular}[c]{@{}l@{}}Why do you think the fennec fox has become an\\ important cultural symbol in Algeria and is even used in\\ sports?\end{tabular}                                                                    \\ \midrule
\multirow{2}{*}{\adjustbox{valign=m}{\includegraphics[width=2.5cm]{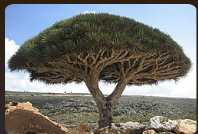}}}                        & \multirow{2}{*}{Role Playing}            & Caption  & \begin{tabular}[c]{@{}r@{}}\RL{ تظهر الصورة واحدة من أشهر وأندر الأشجار في العالم، وهي }\\ \RL{ شجرة دم الأخوين، التي تنمو في جزيرة سقطرى اليمنية. }\end{tabular}                    & \begin{tabular}[c]{@{}l@{}}The picture shows one of the world’s most famous and\\ rarest trees: the Dragon’s Blood Tree, which grows on Yemen’s\\ Socotra Island.\end{tabular}                                                \\  \cmidrule{3-5}
                                          &                                          & Question & \begin{tabular}[c]{@{}r@{}}\RL{ تخيل أنك دليل سياحي في جزيرة سقطرى، كيف تشرح للزوار }\\ \RL{ أهمية شجرة دم الأخوين؟ }\end{tabular}                                                   & \begin{tabular}[c]{@{}l@{}}Imagine you are a tour guide on Socotra—how would you\\ explain to visitors the significance of the Dragon’s Blood Tree?\end{tabular}                                                              \\ \midrule
\multirow{2}{*}{\adjustbox{valign=m}{\includegraphics[width=2.5cm]{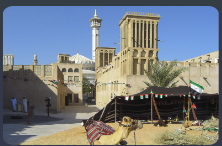}}}                        & \multirow{2}{*}{Problem Solving}         & Caption  & \begin{tabular}[c]{@{}r@{}}\RL{ تظهر الصورة حي البستكية التاريخي في إمارة دبي، حيث تظهر }\\ \RL{ مباني تقليدية ذات تصميم معماري فريد مع أبراج الرياح ... }\end{tabular}               & \begin{tabular}[c]{@{}l@{}}The image shows Dubai’s historic Al-Bastakiya quarter, where traditional buildings\\ with a distinctive architectural style and wind towers can be\\ seen…\end{tabular}                            \\  \cmidrule{3-5}
                                          &                                          & Question & \begin{tabular}[c]{@{}r@{}}\RL{ كيف يمكن الحفاظ على المباني التاريخية التي تظهر في الصورة }\\ \RL{ مع التطور الحضري السريع في المنطقة؟ }\end{tabular}                                 & \begin{tabular}[c]{@{}l@{}}How can the historic buildings shown in the picture be\\ preserved amid the region’s rapid urban development?\end{tabular}                                                                         \\ \midrule
\multirow{2}{*}{\adjustbox{valign=m}{\includegraphics[width=2.5cm]{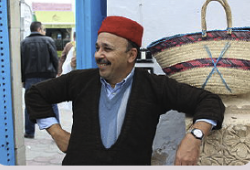}}}                        & \multirow{2}{*}{Modern Context}          & Caption  & \begin{tabular}[c]{@{}r@{}}\RL{ تظهر الصورة رجل تونسي يرتدي لباسا تقليديا يعرف بالكدرون، وهو }\\ \RL{ قطعة ملابس مصنوعة .... }\end{tabular}                                     & \begin{tabular}[c]{@{}l@{}}The photo shows a Tunisian man wearing a traditional garment\\ known as the qadrūn, a piece of clothing made of…\end{tabular}                                                                      \\ \cmidrule{3-5}
                                          &                                          & Question & \begin{tabular}[c]{@{}r@{}}\RL{ كيف يمكن دمج الزي التقليدي الذي يظهر في الصورة في }\\ \RL{ الحياة اليومية الحديثة في تونس؟ }\end{tabular}                                             & \begin{tabular}[c]{@{}l@{}}How can the traditional attire shown in the picture be\\ integrated into modern everyday life in Tunisia?\end{tabular}                                                                             \\ \midrule
\multirow{2}{*}{\adjustbox{valign=m}{\includegraphics[width=2.5cm]{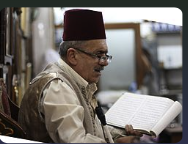}}}                        & \multirow{2}{*}{Original Identification} & Caption  & \begin{tabular}[c]{@{}r@{}}\RL{ تظهر الصورة حكواتيا معاصرا في أحد المقاهي الدمشقية التقليدية وهو }\\ \RL{ يرتدي الطربوش واللباس الشعبي ويجلس .... }\end{tabular}                 & \begin{tabular}[c]{@{}l@{}}The picture shows a contemporary storyteller in a traditional\\ Damascene cafe, wearing a fez and folk dress, seated …\end{tabular}                                                     \\  \cmidrule{3-5}
                                          &                                          & Question & \begin{tabular}[c]{@{}r@{}}\RL{ ما هو الأصل التاريخي أو الجغرافي لشخصية الحكواتي الذي يظهر }\\ \RL{ في الصورة؟ }\end{tabular}                                                         & \begin{tabular}[c]{@{}l@{}}What is the historical or geographical origin of the hakawati\\ figure shown in the picture?\end{tabular}                                                                                          \\ \midrule
\multirow{2}{*}{\adjustbox{valign=m}{\includegraphics[width=2.5cm]{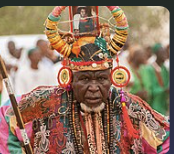}}}                        & \multirow{2}{*}{Scenario Completion}     & Caption  & \begin{tabular}[c]{@{}r@{}}\RL{ تظهر الصورة أحد المشاركين في رقصات الطرق الصوفية في السودان، }\\ \RL{ وهو يرتدي زيا تقليديا مميزا غنيا بالألوان والرموز...... }\end{tabular}     & \begin{tabular}[c]{@{}l@{}}The image depicts a participant in Sudanese Sufi order dances,\\ dressed in a distinctive traditional costume rich in colours and\\ symbols…\end{tabular}                                          \\ \cmidrule{3-5}
                                          &                                          & Question & \begin{tabular}[c]{@{}r@{}}\RL{ بدأ الرجل في الصورة يرقص مع جماعته في احتفال صوفي. }\\ \RL{ اذا توقف توقف عن الرقص في الحفل }\\ \RL{ برأيك، ماذا سيحدث بعد ذلك؟ }\end{tabular}       & \begin{tabular}[c]{@{}l@{}}“The man in the picture began dancing with his group\\ at a Sufi celebration. If he were to stop dancing\\ during the ceremony, what do you think would happen next?”\end{tabular}                 \\ \midrule
\multirow{2}{*}{\adjustbox{valign=m}{\includegraphics[width=2.5cm]{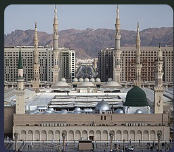}}}                        & \multirow{2}{*}{General QA}             & Caption  & \begin{tabular}[c]{@{}r@{}}\RL{ تظهر الصورة المسجد النبوي في المدينة المنورة، وهو معلم ديني }\\ \RL{ وثقافي بارز في المملكة العربية السعودية. يظهر ف........ }\end{tabular}           & \begin{tabular}[c]{@{}l@{}}The picture shows the Prophet’s Mosque in Medina, a major\\ religious and cultural landmark in the Kingdom of Saudi Arabia.\\ It features …\end{tabular}                                           \\ \cmidrule{3-5}
                                          &                                          & Question & \begin{tabular}[c]{@{}r@{}}\RL{ ما هي الميزة المعمارية البارزة الذي يظهر في الصورة والتي }\\ \RL{ تغطي قبر النبي محمد صلى الله عليه وسلم؟ }\end{tabular}                              & \begin{tabular}[c]{@{}l@{}}What is the prominent architectural feature shown in the picture\\ that covers the grave of the Prophet Muhammad (peace be\\ upon him)?\end{tabular}                                               \\ \midrule
\multirow{2}{*}{\adjustbox{valign=m}{\includegraphics[width=2.5cm]{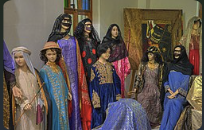}}}                        & \multirow{2}{*}{Modern Context}          & Caption  & \begin{tabular}[c]{@{}r@{}}\RL{ تظهر الصورة مجموعة من الدمى ترتدي الزي التقليدي للمرأة القطرية، }\\ \RL{ وقد عرضت داخل أحد أروقة متحف الشيخ فيصل بن ...... }\end{tabular}           & \begin{tabular}[c]{@{}l@{}}The picture displays a collection of dolls wearing traditional Qatari\\ women’s attire, exhibited in one of the galleries of the\\ Sheikh Faisal bin … Museum.\end{tabular}                        \\ \cmidrule{3-5}
                                          &                                          & Question & \begin{tabular}[c]{@{}r@{}}\RL{ كيف يمكن استخدام الأزياء التقليدية القطرية المعروضة في الصورة في }\\ \RL{ عالم اليوم؟ }\end{tabular}                                                  & \begin{tabular}[c]{@{}l@{}}How can the traditional Qatari costumes displayed in the picture\\ be used in today’s world?\end{tabular}                                                                                          \\ \midrule
\multirow{2}{*}{\adjustbox{valign=m}{\includegraphics[width=2.8cm]{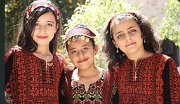}}}                       & \multirow{2}{*}{Prespective Shifting}    & Caption  & \RL{ تظهر الصورة أطفال فلسطينيين يرتدون الزي الفلسطيني التقليدي. }                                                                                                                    & The picture shows Palestinian children dressed in traditional Palestinian clothing.                                                                                                                                           \\ \cmidrule{3-5}
                                          &                                          & Question & \begin{tabular}[c]{@{}r@{}}\RL{ ناقش كيف يختلف فهم الزي الفلسطيني التقليدي الذي في الصورة }\\ \RL{ ودلالته عند النظر إليه من وجهة نظر فلسطيني يعيش في }\\ \RL{ المهجر؟ }\end{tabular} & \begin{tabular}[c]{@{}l@{}}Discuss how the understanding and symbolism of the traditional Palestinian\\ attire in the picture differ when viewed from the perspective\\ of a Palestinian living in the diaspora.\end{tabular} \\ \midrule
\multirow{2}{*}{\adjustbox{valign=m}{\includegraphics[width=1.8cm]{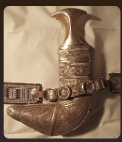}}}                       & \multirow{2}{*}{Chronological Sequence}  & Caption  & \begin{tabular}[c]{@{}r@{}}\RL{ تظهر الصورة الخنجر العماني التقليدي، والمعروف أيضا باسم “الجنبية” في }\\ \RL{ بعض المناطق. يتميز الخنجر ... }\end{tabular}                       & \begin{tabular}[c]{@{}l@{}}The picture depicts the traditional Omani dagger, also known as\\ the janbiyyah in some regions. The dagger is distinguished by\\ …\end{tabular}                                                   \\ \cmidrule{3-5}
                                          &                                          & Question & \RL{ كيف تطور استخدام العنصر الذي يظهر في الصورة عبر الزمن؟ }                                                                                                                        & \begin{tabular}[c]{@{}l@{}}How has the use of the item shown in the\\ picture evolved over time?\end{tabular}                                                                                                                 \\ \midrule
\multirow{2}{*}{\adjustbox{valign=m}{\includegraphics[width=2.5cm]{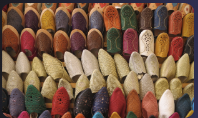}}}                       & \multirow{2}{*}{Comparative analysis}    & Caption  & \begin{tabular}[c]{@{}r@{}}\RL{ تظهر الصورة مجموعة متنوعة من الأحذية التقليدية المغربية الملونة والمزخرفة، }\\ \RL{ مرتبة بشكل منظم .... }\end{tabular}                              & \begin{tabular}[c]{@{}l@{}}The image shows a neatly arranged selection of colourful, ornamented\\ traditional Moroccan shoes …\end{tabular}                                                                                   \\ \cmidrule{3-5}
                                          &                                          & Question & \begin{tabular}[c]{@{}r@{}}\RL{ كيف يختلف الحذاء التقليدي الذي يظهر في الصورة عن الأحذية }\\ \RL{ الحديثة من حيث التصميم والوظيفة؟ }\end{tabular}                                     & \begin{tabular}[c]{@{}l@{}}How does the traditional shoe shown in the picture differ\\ from modern footwear in terms of design and function?\end{tabular}                                                                     \\ \midrule
\multirow{2}{*}{\adjustbox{valign=m}{\includegraphics[width=2.5cm]{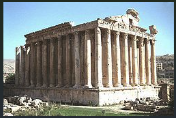}}}                       & \multirow{2}{*}{Original Identification} & Caption  & \begin{tabular}[c]{@{}r@{}}\RL{ تظهر الصورة أطلال معبد روماني ضخم في مدينة بعلبك اللبنانية، }\\ \RL{ وهو أحد أبرز المعالم الأثرية في لبنان والشرق الأوسط. }\end{tabular}             & \begin{tabular}[c]{@{}l@{}}The picture presents the ruins of a massive Roman temple\\ in Baalbek, Lebanon—one of the most prominent archaeological landmarks in\\ Lebanon and the Middle East.\end{tabular}                   \\ \cmidrule{3-5}
                                          &                                          & Question & \RL{ ما هو الأصل التاريخي للعنصر المعماري الموجود في الصورة؟ }                                                                                                                        & \begin{tabular}[c]{@{}l@{}}What is the historical origin of the architectural element shown\\ in the picture?\end{tabular}                                                                                                    \\ \midrule
\multirow{2}{*}{\adjustbox{valign=m}{\includegraphics[width=2.5cm]{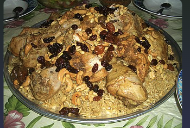}}}                       & \multirow{2}{*}{Original Identification} & Caption  & \begin{tabular}[c]{@{}r@{}}\RL{ تظهر الصورة طبقا شهيا من المجبوس أو الكبسة الخليجية، وهو }\\ \RL{ أحد أشهر ال }\end{tabular}                                                       & \begin{tabular}[c]{@{}l@{}}The picture shows a mouth-watering dish of Gulf-style majboos (also\\ called kabsa), one of the region’s most celebrated rice dishes.\end{tabular}                                                 \\ \cmidrule{3-5}
                                          &                                          & Question & \begin{tabular}[c]{@{}r@{}}\RL{ كيف يختلف طبق الذي في الصورة عن أطباق الأرز التقليدية }\\ \RL{ في دول عربية أخرى مثل المنسف الأردني؟ }\end{tabular}                                   & \begin{tabular}[c]{@{}l@{}}How does the dish in the picture differ from other\\ traditional rice dishes in Arab countries, such as Jordanian mansaf?\end{tabular}                                                             \\ \midrule
\multirow{2}{*}{\adjustbox{valign=m}{\includegraphics[width=1.7cm]{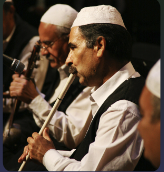}}}                       & \multirow{2}{*}{Cause and Effect}        & Caption  & \begin{tabular}[c]{@{}r@{}}\RL{ تظهر الصورة مجموعة من العازفين العراقيين من فرقة فنون شعبية، }\\ \RL{ يرتدون الزي التقليدي، ويؤدون عرضا موسيقيا .. }\end{tabular}                  & \begin{tabular}[c]{@{}l@{}}The image features a group of Iraqi musicians from a\\ folk-arts ensemble, dressed in traditional attire and giving a musical\\ performance …\end{tabular}                                         \\ \cmidrule{3-5}
                                          &                                          & Question & \RL{ لماذا يستخدم المزمار في الموسيقى الشعبية العراقية؟ }                                                                                                                            & Why is the mizmar used in Iraqi folk music?                                                                                                                                                                                   \\ \midrule
\multirow{2}{*}{\adjustbox{valign=m}{\includegraphics[width=2.5cm]{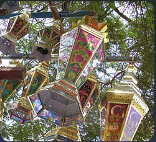}}}                       & \multirow{2}{*}{Original Identification} & Caption  & \begin{tabular}[c]{@{}r@{}}\RL{ الصورة تظهر مجموعة من فوانيس رمضان التقليدية المعلقة على أغصان }\\ \RL{ شجرة، بألوان زاهية وزخارف مبهجة....... }\end{tabular}                       & \begin{tabular}[c]{@{}l@{}}The photo shows a collection of traditional Ramadan lanterns hanging\\ from tree branches, glowing in vivid colours and cheerful designs\\ …\end{tabular}                                          \\ \cmidrule{3-5}
                                          &                                          & Question & \begin{tabular}[c]{@{}r@{}}\RL{ ما هو الأصل التاريخي أو الجغرافي للعنصر الثقافي الموجود في }\\ \RL{ الصورة؟ }\end{tabular}                                                            & \begin{tabular}[c]{@{}l@{}}What is the historical or geographical origin of the cultural\\ element shown in the picture?\end{tabular}    \\ \bottomrule                                                                                     
                                     
\end{tabular}%
}
\caption{Provides English version of the main figure, including the image, question types, and translations of the original Arabic texts}
\label{appdx_table:examples}
\end{table*}

\begin{figure*}[ht]  
\centering
\begin{tcolorbox}[
    title={\bfseries Prompt},
    width=\textwidth,
    colframe=gray!75!black,
    colback=blue!5,
    colbacktitle=gray!75!black,
    coltitle=white,
    boxsep=3pt,
    arc=1mm,
    fontupper=\small\ttfamily,
    before upper={\setlength{\parskip}{0.5em}\setlength{\baselineskip}{0.9\baselineskip}},
]
{\color{gray!80!black}\bfseries We are building Arabic culture multimodal datasets. In some cases, a single cultural concept is shared across multiple Arabic-speaking countries but manifests in distinct ways—visually, in preparation, or in usage. For instance, Kabsa is a traditional dish enjoyed in places like Saudi Arabia, Yemen, and Qatar, yet each locale has its own way of preparing and seasoning it. Likewise, “Agal” is a type of traditional head accessory worn across various Arab regions, though its style or method of wearing can differ from one country to another. These nuanced variations across shared concepts highlight the rich cultural diversity that exists within the broader Arabic-speaking world. We called this “Shared Concepts”. We introduce Pearl-Shared Concepts as a specialized benchmark that evaluates the LVLM’s ability to generalize and reason consistently about cultural concepts shared across multiple Arabic-speaking regions. Initially, we identified culturally common concepts and curated diverse multimodal examples (images and texts) reflecting variations across regions. We then developed template-based multiple-choice and true/false questions to systematically assess how robustly LVLMs generalize shared concepts despite variations in visual representation or cultural context. I will provide you with some images and templates of questions (MCQ and true and false), and I want you to generate questions based on the templates. Be creative and don’t stick to these templates. Some of the templates require one image question and the others require multiple images. I will provide you with the name of the shared concept. The images are numbered in the format of {number}\_{name of country}.{extension}. Also, I will provide you with a shared concept name. I want you to generate 2-3 single-image questions, 2-3 multiple-image questions. }

{\color{blue!50!black}\bfseries The output format should be in the following json format:}

\begin{tcolorbox}[colback=gray!10, colframe=gray!40, size=fbox, boxrule=0.5pt, left=1pt, right=1pt, top=1pt, bottom=1pt]
\footnotesize
\ttfamily
\{
  “ID”: 1, \\
  “Question”: Question? (In Arabic), \\
  “Choices”: \{ \\
    “A”: Choice 1 (In Arabic), \\
    “B”: Choice 2 (In Arabic), \\
    “C”: Choice 3 (In Arabic), \\
    “D”: Choice 3 (In Arabic) \\
  \},\\
  “Answer”: “A”,\\
  “Single\_Multiple”: “Multiple”,\\
  “Selected\_images”: [1, 2, 3, 4]\\
\}
\end{tcolorbox}

The concept name: Concept Name in Arabic: \{concept\_name\}

\begin{tcolorbox}[colback=gray!10, colframe=gray!40, size=fbox, boxrule=0.5pt, left=1pt, right=1pt, top=1pt, bottom=1pt]
\footnotesize
\ttfamily
\{Template-Based Questions (In Arabic)\\
What is another name for [element X] in this country?\\
What is a key ingredient that distinguishes the preparation of [dish X] here? \\
On what occasion is the depicted [element X] worn or presented?\\
True or False: [Dish X] is usually served with Arabic bread in this country.\\
True or False: The visible color on [element X] indicates its Qatari origin.\\
Arrange the images according to the level of spiciness from mildest to spiciest.\\
Which of these images represents the oldest historical form of [element X]?\\
True or False: All the images show [element X] being worn in a similar way."\}
\end{tcolorbox}

\end{tcolorbox}
\caption{Illustrates the prompt used to generate Q\&A pairs for~\pearlx using \texttt{GPT-o3}. in the prompt, we include few-shot examples based on templates for both single-image and multi-image question types. }
\label{appdx_fig:prompt_new}
\end{figure*}

\begin{figure*}[!ht]  
\centering
\begin{tcolorbox}[
    title={\bfseries Prompt},
    width=\textwidth,
    colframe=gray!75!black,
    colback=blue!5,
    colbacktitle=gray!75!black,
    coltitle=white,
    boxsep=3pt,
    arc=1mm,
    fontupper=\small\ttfamily,
    before upper={\setlength{\parskip}{0.5em}\setlength{\baselineskip}{0.9\baselineskip}},
]
{\color{blue!80!black}\bfseries multiple\_choice's promprt}

\begin{tcolorbox}[colback=gray!10, colframe=gray!40, size=fbox, boxrule=0.5pt, left=1pt, right=1pt, top=1pt, bottom=1pt]
\footnotesize
\ttfamily
You are an impartial evaluator.\\
                        
TASK: Decide if the candidate's choice is correct **ignoring surface form** (letter, synonym, capitalisation).  Return ONLY "1" for correct, "0" for incorrect.\\
                        
 \textbf{Question with options:}\\
                        \{question\}\\
                        \textbf{Gold correct answer:}\\
                        \{ground\_truth\}\\
                        \textbf{Candidate's chosen answer:}\\
                        \{predicted\_answer\}\\
                        
                        Reply with 1 or 0 – nothing else.
\end{tcolorbox}

{\color{blue!80!black}\bfseries true\_false's promprt}

\begin{tcolorbox}[colback=gray!10, colframe=gray!40, size=fbox, boxrule=0.5pt, left=1pt, right=1pt, top=1pt, bottom=1pt]
\footnotesize
\ttfamily
You are an impartial evaluator.\\
                        
                        TASK: Compare the candidate's short answer with the gold answer.
                        If they express the **same fact** (allowing synonyms, paraphrase, spelling variants) return "1".
                        Otherwise return "0".
                        Give no explanation.\\
                        \textbf{Statement:}\\
                        \{question\}\\
                        \textbf{Gold label (True/False):}\\
                        \{ground\_truth\}\\
                        \textbf{Candidate label:}\\
                        \{predicted\_answer\}\\
                        
                        Reply with 1 or 0 – nothing else.
\end{tcolorbox}

{\color{blue!80!black}\bfseries open-ended's promprt}

\begin{tcolorbox}[colback=gray!10, colframe=gray!40, size=fbox, boxrule=0.5pt, left=1pt, right=1pt, top=1pt, bottom=1pt]
\footnotesize
\ttfamily
You are an **impartial multimodal evaluator** for Arabic cultural benchmark. \\ 
        Your task is to grade the candidate’s Arabic answer to a culture‑focused question,
        using **both** the reference text and the attached image.\\
        ***Output ONE JSON object only*** in this schema:\\
        \{\{
          "correctness": <0‑5>,
          "coherence":  <0‑5>,
          "detail":     <0‑5>,
          "fluency":    <0‑5>
       \}\}\\
        \#\#\# Scoring rubric (0= very poor,5= excellent) \\
        • **Correctness 40\%\*\* – factual agreement with the gold reference *and* visual evidence.\\  
        • **Coherence 20\%**  – logical, no contradictions, clear progression.\\  
        • **Detail 20\%**     – culturally specific names, dates, artefacts, dialect terms (text or image).\\  
        • **Fluency 20\%**    – smooth Arabic (or requested dialect), correct grammar \& style.\\
  
        \textbf{Image description:}\\ \{image\_description\}\\
        \textbf{Question:} \{question\}\\
        \textbf{Gold reference answer:}\\ \{ground\_truth\}\\
        \textbf{Candidate answer:}\\  \{predicted\_answer\}\\
        
        Respond with the JSON object only— **no additional text**

\end{tcolorbox}

\end{tcolorbox}
\caption{Evaluation prompts used for automatic judgment across different question formats in the Pearl benchmark. The \texttt{multiple\_choice} prompt evaluates whether the predicted choice matches the correct answer, allowing for variation in surface form (e.g., casing or synonyms). The \texttt{true\_false} prompt compares semantic equivalence between gold and predicted binary labels. The \texttt{short\_answer} prompt assesses alignment between candidate and gold judgments on factual statements. All prompts instruct the model to return strictly binary outcomes (1 or 0) with no explanation. Detailed evaluation prompts are available on the project GitHub repository: \href{https://github.com/UBC-NLP/pearl}{https://github.com/UBC-NLP/pearl}.}

\label{appdx_fig:prompt_eval}
\end{figure*}
\clearpage
\begin{table*}[!ht]
\centering
\resizebox{\textwidth}{!}{%
\begin{tabular}{lllllllll}
\toprule
\rowcolor{gray!30}
\textbf{Architecture} & \textbf{Clothes} & \textbf{Fauna} & \textbf{Flora} & \textbf{Food} & \textbf{Geography} & \textbf{Handicrafts} & \textbf{Landmarks} & \textbf{Music} \\
\midrule
Courtyard Houses & Keffiyeh \& Agal & Camel & Date Palm & Couscous & Nile River Valley & Al-Sadu Weaving & Historic Citadels and Forts & Oud \\

Souk Bazaars & Thawb/Dishdasha & Falconry & Olive Tree & Kabsa/Machbūs & Mediterranean Coast & Carpet and Rug Weaving & Grand Mosques of Early Islam & Dabke \\

Wind Catchers & Abaya & Saluki Dogs & Cedar Tree & Mandi & Atlas Mountains & Embroidery (Tatreez) & Historic City ''Old Towns'' (Medinas) & Mizmar \\

Mudbrick Architecture & Jellabiya & Arabian Oryx & Argan Tree & Falafel & Red Sea Coast & Khanjar & Caravanserais (Khans) & Sword Dance (Al-'Ardha) \\

Historic Citadels and Forts & Kaftan & Fennec Fox & Jasmine & Hummus & Desert Oases & & & \\

Grand Mosques of Early Islam & Keffiyeh & & & Ful Medames & & & & \\

Historic City ``Old Towns'' (Medinas) & Djellaba & & & Stuffed Vegetables (Mahshi/Dolma) & & & & \\

Caravanserais (Khans) & Bisht & & & Shawarma & & & & \\

& Fez (Tarboosh) & & & Arabic Coffee & & & & \\

& & & & Harissa & & & & \\

& & & & Mulukhiyah & & & & \\

& & & & Shakshouka & & & & \\

& & & & Asida & & & & \\

& & & & Qatayef & & & & \\
\bottomrule
\end{tabular}%
}
\caption{Shows \textit{PearlX}'s comprehensive list of $61$ culturally diverse concepts spanning various categories across Arab countries, which were used to generate Q\&A pairs for benchmarking}
\label{tab:shared_concepts_elemnts}
\end{table*}

\begin{table*}[!ht]
\centering
\resizebox{\textwidth}{!}{%
\begin{tabular}{crrl}
\toprule
\textbf{Type}                              & \multicolumn{1}{c}{\textbf{Template}}                                                                                                 & \multicolumn{1}{c}{\textbf{Example}}                                                                                                                                                                                 & \multicolumn{1}{c}{\textbf{Q-Type}} \\ \midrule
\multirow{10}{*}{\rotatebox{90}{Multiple Image Templates}} & \RL{ طابق كل صورة بالدولة الصحيحة استناداً إلى اختلاف نمط [العنصر ]. }                                                               & \begin{tabular}[c]{@{}r@{}}\RL{ طابق كل صورة بالعقال القطري، السعودي، العراقي. }\\  \RL{ ⓵ ( القطري، السعودي، العراقي. ⓶ القطري، السعودي، العراقي ⓷ القطري، }\\ \RL{ السعودي، العراقي. }\end{tabular} & Choose                              \\  \cmidrule{2-4}
                                           & \RL{ صنِّف الصور إلى مجموعتين بحسب نوع البهارات المستخدمة في [الطبق ]. }                                                             & \begin{tabular}[c]{@{}r@{}}\RL{ صنّف صور الكبسة إلى مجموعة تستخدم الهيل وأخرى لا تستخدمه. }\\ \RL{  }\end{tabular}                                                                                                   & Categorieze                         \\  \cmidrule{2-4}
                                           & \RL{ اختر الصورة التي لا تنتمي إلى نفس المفهوم المشترك. }                                                                             & \RL{ اختر الصورة التي لا تمثل طبق كبسة بين الصور الأربع. }                                                                                                                                                           & Choose                              \\  \cmidrule{2-4}
                                           & \RL{ رتِّب الصور حسب درجة حدة التوابل من الأخف إلى الأشد. }                                                                           & \RL{ رتّب صور أطباق الكبسة السعودية، العمانية، واليمنية حسب حدة التوابل. }                                                                                                                                           & Reorder                             \\  \cmidrule{2-4}
                                           & \RL{ أي من هذه الصور تُمثِّل أقدم شكل تاريخي لـ[العنصر ]؟ }                                                                          & \RL{ أي من هذه الصور تمثّل الشكل التاريخي الأقدم للعقال؟ }                                                                                                                                                           & Idenify                             \\  \cmidrule{2-4}
                                           & \RL{ صحيح أم خطأ: كل الصور تُظهر [العنصر ] يُرتدى بطريقة متشابهة. }                                                                  & \RL{ صحيح أم خطأ: كل الصور تُظهر العقال يُرتدى بالطريقة نفسها. }                                                                                                                                                     & T/F                                 \\  \cmidrule{2-4}
                                           & \RL{ اختر الصورة التي توضّح خطوة الطهي الأولى في إعداد [الطبق ]. }                                                                   & \begin{tabular}[c]{@{}r@{}}\RL{ اختر الصورة التي توضّح أول خطوة في طهي الكبسة بين الصور }\\ \RL{ الثلاث. }\end{tabular}                                                                                              & MCQ                                 \\  \cmidrule{2-4}
                                           & \RL{ أي من هذه الصور تُمثل نسخة حديثة مطوَّرة من [العنصر ]؟ }                                                                        & \begin{tabular}[c]{@{}r@{}}\RL{ أي من هذه الصور تمثل نسخه حديثة مطورة للعقال باستخدام خامات }\\ \RL{ صناعية؟ }\end{tabular}                                                                                          & Choose                              \\  \cmidrule{2-4}
                                           & \RL{ طابق كل صورة بالدولة الصحيحة استناداً إلى اختلاف نمط [العنصر ]. }                                                               & \RL{ طابق صور العقال مع الدول: أ) قطر ب) السعودية ج) العراق }                                                                                                                                                        & MCQ                                 \\  \midrule
\multirow{10}{*}{\rotatebox{90}{Single Image Templates}} & \RL{ ما الاسم الآخر الذي يُطلق على [العنصر ] في دولة؟ }                                                                          & \RL{ ما الاسم الآخر الذي يُطلق على «كبسة» في اليمن؟ }                                                                                                                                                                & T/F                                 \\  \cmidrule{2-4}
                                           & \begin{tabular}[c]{@{}r@{}}\RL{ اذكر المكوّن الرئيس الذي يميز [الطبق] في هذه الدولة عن }\\ \RL{ نسخته في دولةٍ أخرى. }\end{tabular} & \RL{ اذكر المكوّن الرئيس الذي يميز كبسة سلطنة عُمان عن كبسة السعودية. }                                                                                                                                              & T/F                                 \\  \cmidrule{2-4}
                                           & \RL{ في أي مناسبة أو فصل عادةً ما يُستخدم/يُرتدى [العنصر ] هنا؟ }                                                                    & \RL{ في أي مناسبة عادةً ما يُرتدى هذا العقال في العراق؟ }                                                                                                                                                            & T/F                                 \\  \cmidrule{2-4}
                                           & \begin{tabular}[c]{@{}r@{}}\RL{ ما اللون/الزخرفة التي تظهر في [العنصر ] وتدلّ على أصلِه من }\\ \RL{ هذه الدولة؟ }\end{tabular}       & \RL{ ما الزخرفة التي تظهر في هذا العقال وتدلّ على أصله القطري؟ }                                                                                                                                                     & T/F                                 \\  \cmidrule{2-4}
                                           & \begin{tabular}[c]{@{}r@{}}\RL{ صحيح أم خطأ: يُقدَّم [الطبق ] عادةً مع الخبز العربي في }\\ \RL{ هذه الدولة. }\end{tabular}           & \RL{ صحيح أم خطأ: يُقدَّم الفول المدمس مع الخبز العربي في السودان. }                                                                                                                                                 & T/F                                 \\  \cmidrule{2-4}
                                           & \RL{ ما الاسم الآخر الذي يُطلق على [العنصر ] في هذه الدولة؟ }                                                                        & \begin{tabular}[c]{@{}r@{}}\RL{ ما الاسم الآخر الذي يُطلق على «كبسة» في اليمن؟ }\\ \RL{ أ) المندي ب) السلتة ج) الزربيان د) الصالونة }\end{tabular}                                                                   & MCQ                                 \\  \cmidrule{2-4}
                                           & \RL{ أي مكوّن رئيس يميّز إعداد [الطبق ] هنا؟ }                                                                                       & \begin{tabular}[c]{@{}r@{}}\RL{ أي مكوّن رئيس يميّز كبسة سلطنة عُمان؟ }\\ \RL{ أ) الحلبة ب) الهيل ج) اللومي د) الكركم }\end{tabular}                                                                                 & MCQ                                 \\  \cmidrule{2-4}
                                           & \RL{ في أي مناسبة يُرتدى/يُقدَّم [العنصر ] الموضَّح؟ }                                                                               & \begin{tabular}[c]{@{}r@{}}\RL{ في أي مناسبة يُرتدى هذا العقال العراقي؟ }\\ \RL{ أ) الأعراس ب) العمل اليومي ج) الصلاة د) الحج }\end{tabular}                                                                         & MCQ                                 \\  \cmidrule{2-4}
                                           & \RL{ أي حقبة تاريخية ارتبطت بظهور [العنصر ] في هذه الدولة؟ }                                                                         & \begin{tabular}[c]{@{}r@{}}\RL{ أي حقبة تاريخية ارتبطت بظهور «المجبوس» في الكويت؟ }\\ \RL{ أ) العباسيون ب) العثمانيون ج) الأندلسيون د) الفاطميون }\end{tabular}                                                      & MCQ                                 \\ \cmidrule{2-4}
                                           & \RL{ ما أداة الطهي التقليدية المستخدمة لإعداد [الطبق ] هنا؟ }                                                                        & \begin{tabular}[c]{@{}r@{}}\RL{ ما أداة الطهي التقليدية المستخدمة لإعداد الكبسة السعودية؟ }\\ \RL{ أ) القدر الضغاط ب) التنور ج) القدر المقلوب د) المضغوط }\end{tabular}                                              & MCQ   \\ \bottomrule                             
\end{tabular}%
}
\caption{Examples of question templates for both multiple and single image prompts in Arabic that we used to generate True/False and MCQ questions using \texttt{ChatGPT-o3}}
\label{appdx_tab:templete_examples}
\end{table*}

\clearpage
\section{Fine-Grained Performance Analysis on \pearlt by Country and Question Type}
\label{app:fine_grained_analysis}
\begin{figure*}[!ht]
    \centering
    \includegraphics[width=0.80\linewidth, trim={0 0 0 9}, clip]{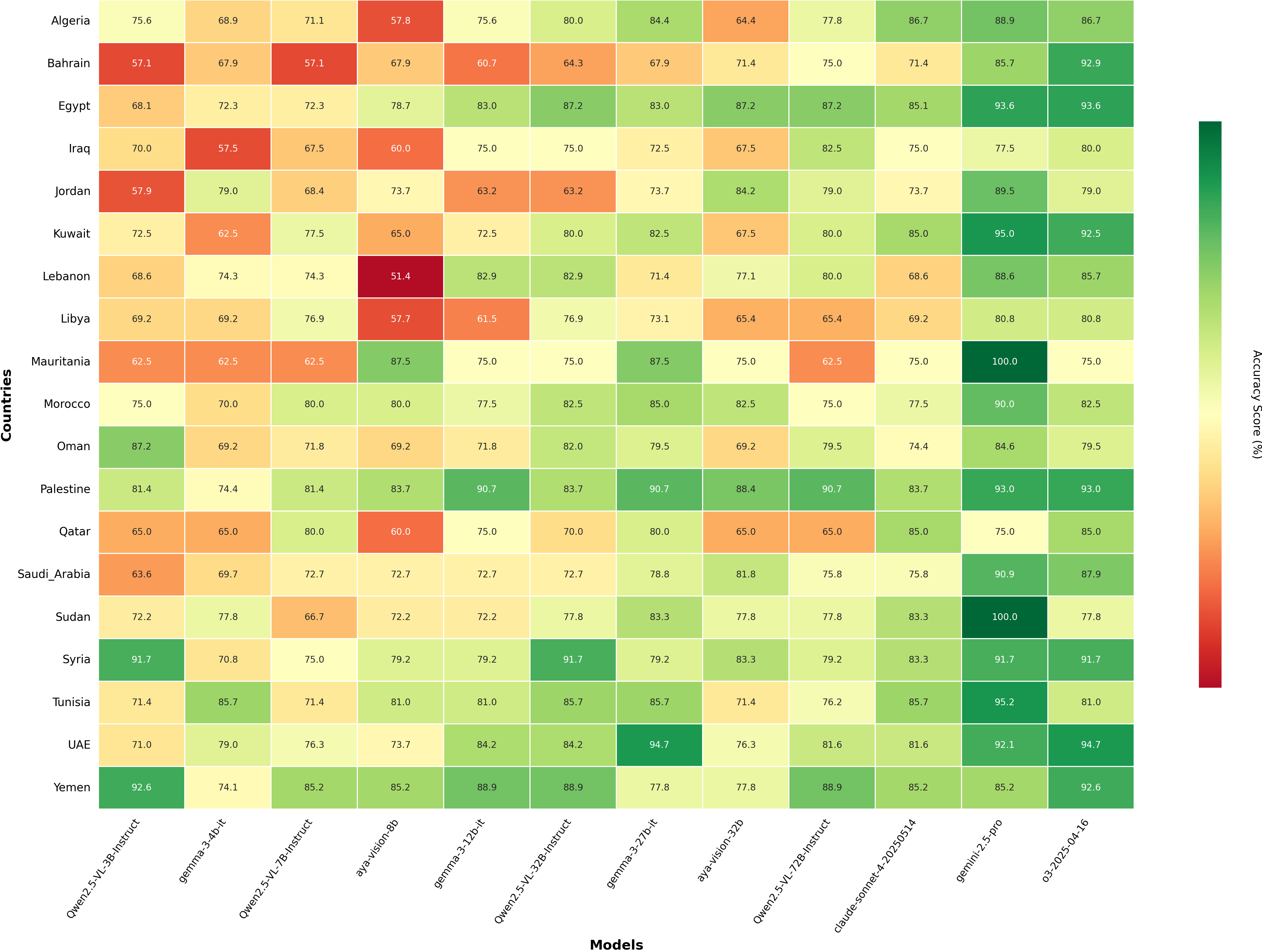}
    \caption{Heatmap of accuracy scores (\%) on closed-form questions, broken down by model and country. }
    \label{fig:accuracy_by_model_country}
\end{figure*}

\noindent
In this section, we provide a more granular analysis of model performance on the \pearlt benchmark. Figure~\ref{fig:accuracy_by_model_country} presents a detailed breakdown of accuracy scores on closed-form questions, categorized by both model and country. The heatmap illustrates that while larger proprietary models like \texttt{Gemini 2.5 Pro} and the \texttt{o3} models consistently achieve high accuracy across nearly all countries, the performance of open-source models varies significantly depending on the geographic context. For instance, several models show lower performance on questions related to Lebanon and Jordan, indicating potential gaps in their regional cultural knowledge for fact-based retrieval tasks.

\begin{figure*}[!ht]
    \centering
    \includegraphics[width=0.80\linewidth, trim={0 0 0 9}, clip]{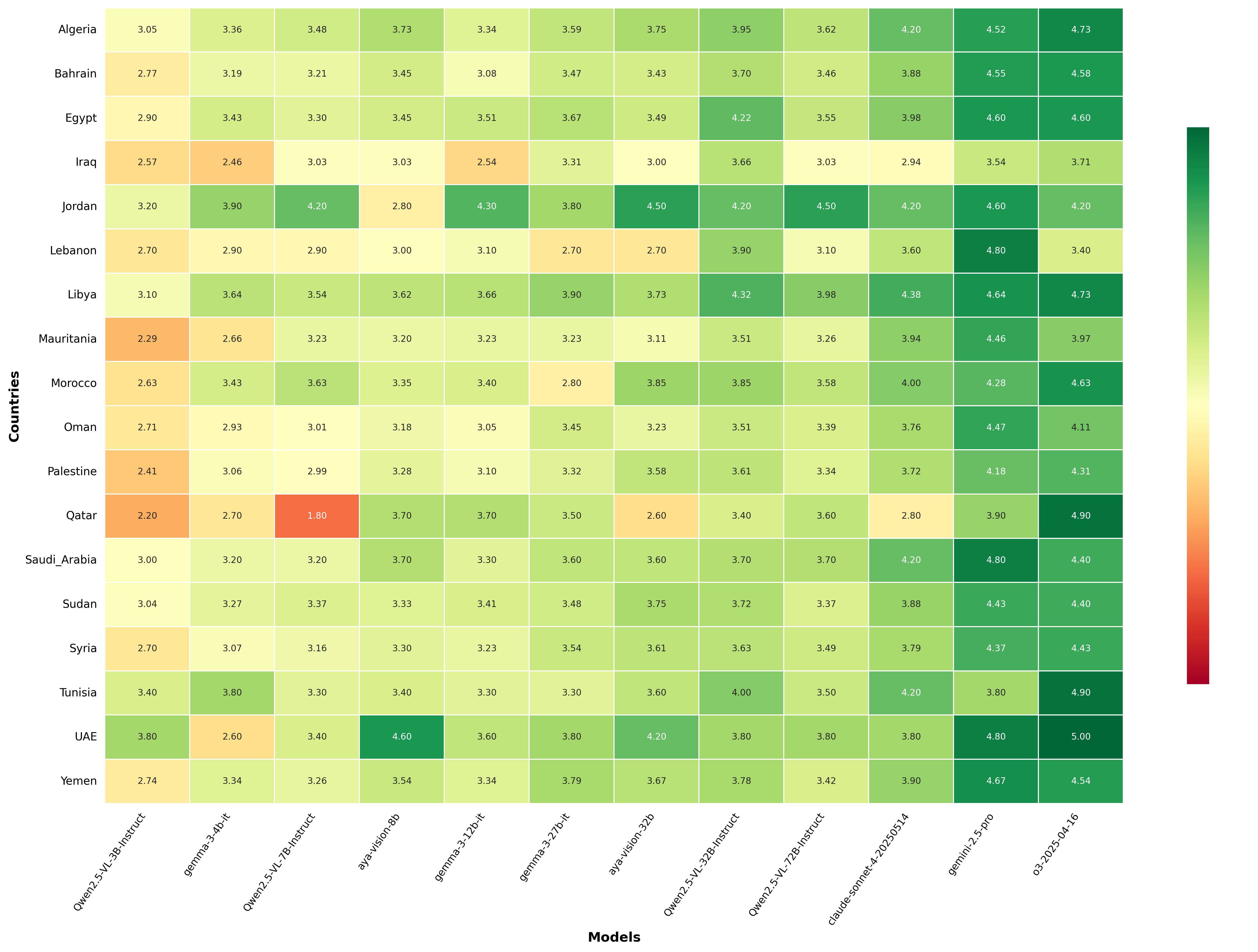}
    \caption{Heatmap of the Overall Score (1-5) for open-ended questions, analyzed by model and country}
    \label{fig:overall_by_model_country}
\end{figure*}

\begin{figure*}[!ht]
    \centering
    \includegraphics[width=0.80\linewidth, trim={0 0 0 9}, clip]{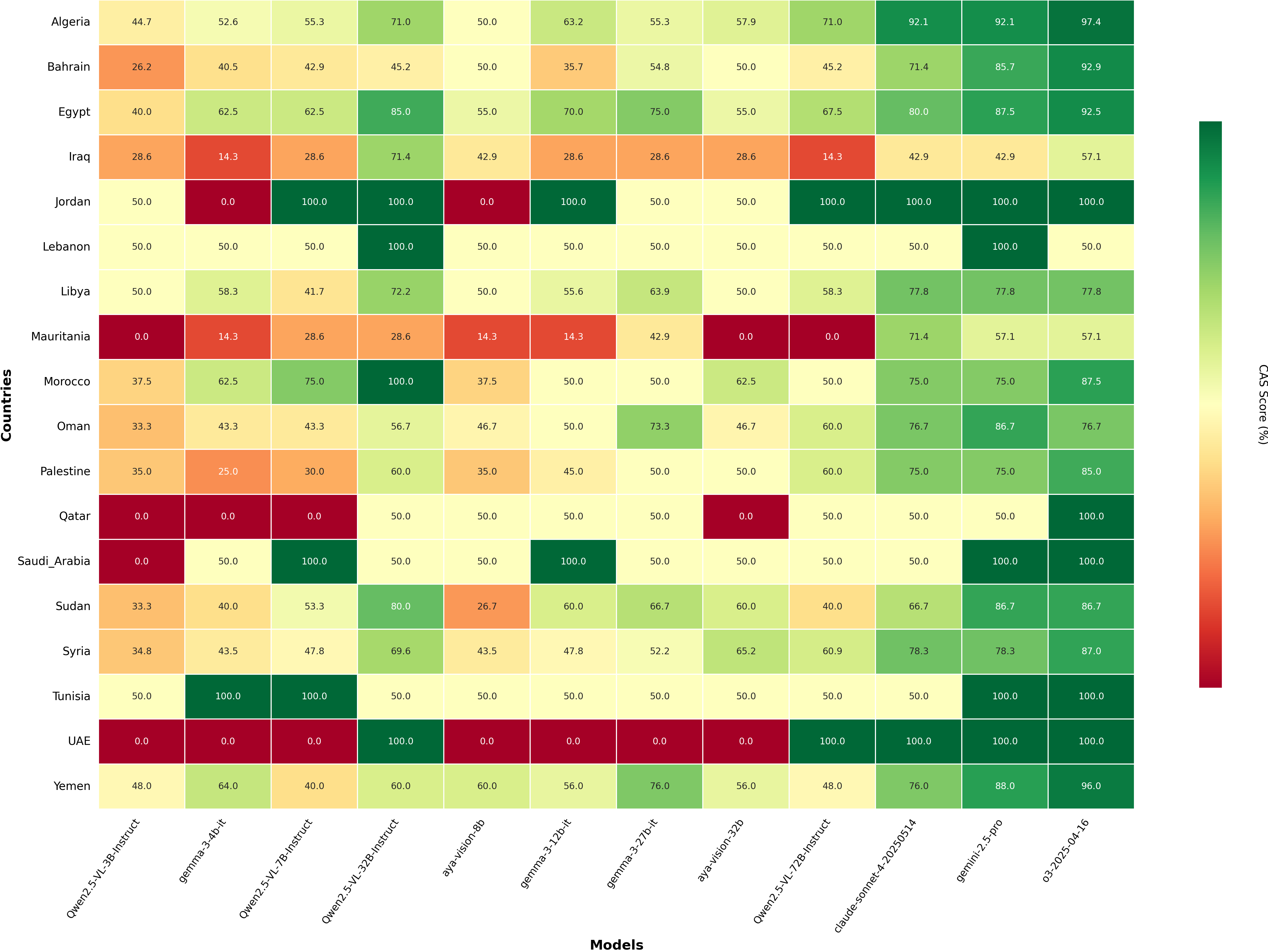}
    \caption{Heatmap of the Cultural Awareness Score (CAS, in \%) for open-ended questions, by model and country}
    \label{fig:cas_by_model_country}
\end{figure*}

For the more challenging open-ended questions, we offer several detailed views. Figures~\ref{fig:overall_by_model_country} and~\ref{fig:cas_by_model_country} display the \textit{Overall score} and \textit{CAS score}, respectively, broken down by model and country. These results underscore the difficulty of generating culturally nuanced text. The proprietary models again lead in performance, but even they show variability, particularly in the CAS metric where scores for countries like Mauritania and Qatar are notably lower for many models. This highlights that a model's ability to be culturally aware is not uniform and can be highly dependent on the specific regional culture being evaluated.

\begin{figure*}[!ht]
    \centering
    \includegraphics[width=0.80\linewidth, trim={0 0 0 9}, clip]{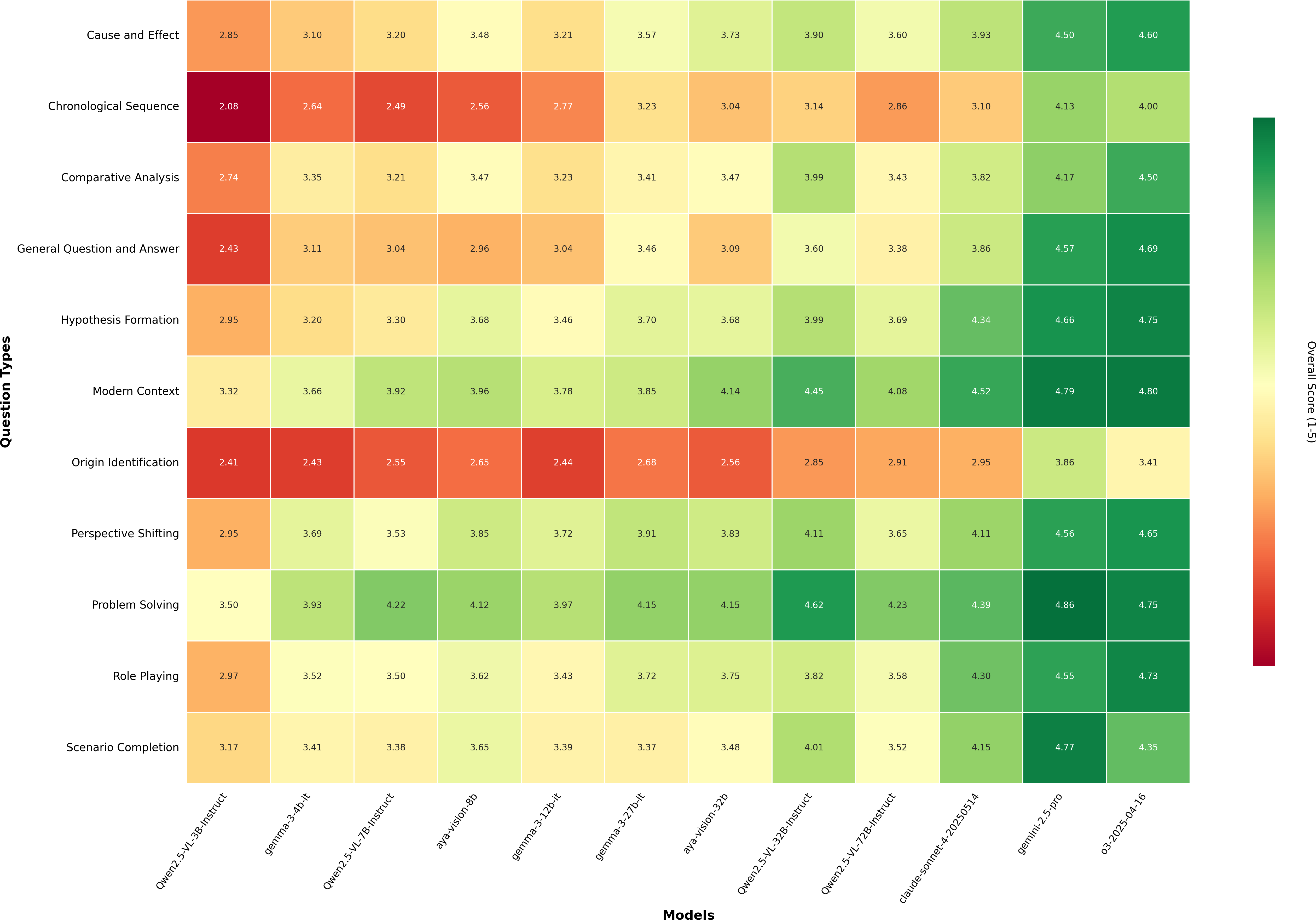}
    \caption{Heatmap of the Overall Score (1-5) for open-ended questions, broken down by model and question type}
    \label{fig:overall_by_type_country}
\end{figure*}

To further dissect the reasoning capabilities of each model, Figure~\ref{fig:overall_by_type_country} provides a performance breakdown of the Overall Score by the 11 different open-ended question types. This analysis reveals specific strengths and weaknesses in model reasoning. Complex tasks such as \textit{Chronological Sequence} and \textit{Origin identification} prove to be challenging for smaller models, which often score poorly. In contrast, more advanced models demonstrate stronger and more consistent performance across these sophisticated reasoning categories, emphasizing our finding that reasoning-centric alignment is crucial for achieving deep cultural comprehension.

\clearpage

\end{document}